\documentclass[letterpaper]{article} 
\usepackage{aaai24}  
\usepackage{times}  
\usepackage{helvet}  
\usepackage{courier}  
\usepackage[hyphens]{url}  
\usepackage{graphicx} 
\usepackage{natbib}  
\usepackage{caption} 
\usepackage[table]{xcolor}
\usepackage{amsmath}
\usepackage{amssymb}
\usepackage{booktabs}
\usepackage{multirow}
\usepackage{makecell}
\usepackage{bm}
\usepackage{enumerate}
\usepackage{arydshln}
\usepackage{booktabs}
\usepackage{multirow}
\usepackage{tabularx}
\usepackage{algorithmic}
\usepackage{algorithm}
\usepackage{amsthm}
\usepackage{color}
\usepackage{xfrac}
\usepackage{threeparttable}
\usepackage{siunitx}
\usepackage{caption}
\usepackage{hhline}
\usepackage{subcaption}
\usepackage{pifont}
\newcommand{\cmark}{\ding{51}}%
\newcommand{\xmark}{\ding{55}}%
\usepackage{algorithm}
\usepackage{algorithmic}

\usepackage{newfloat}
\usepackage{listings}
\DeclareCaptionStyle{ruled}{labelfont=normalfont,labelsep=colon,strut=off} 
\floatstyle{ruled}
\newfloat{listing}{tb}{lst}{}
\floatname{listing}{Listing}
\title{Few-shot Neural Radiance Fields Under Unconstrained Illumination}

\author {
    SeokYeong Lee\textsuperscript{\rm 1,\rm 2},
    JunYong Choi\textsuperscript{\rm 1,\rm 2},
    Seungryong Kim\textsuperscript{\rm 2},
    Ig-Jae Kim\textsuperscript{\rm 1,\rm 3,\rm 4},
    Junghyun Cho\textsuperscript{\rm 1,\rm 3,\rm 4}
}
\affiliations {
    \textsuperscript{\rm 1} Korea Institute of Science and Technology, Seoul \ 
    \textsuperscript{\rm 2} Korea University, Seoul \
    \textsuperscript{\rm 3} AI-Robotics, KIST School, University of Science and Technology \
    \textsuperscript{\rm 4} Yonsei-KIST Convergence Research Institute, Yonsei University \\
    \{shapin94, happily, drjay, jhcho\}@kist.re.kr \  seungryong\_kim@korea.ac.kr \\
}

\begin{document}

\maketitle

\begin{abstract}
In this paper, we introduce a new challenge for synthesizing novel view images in practical environments with limited input multi-view images and varying lighting conditions. Neural radiance fields (NeRF), one of the pioneering works for this task, demand an extensive set of multi-view images taken under constrained illumination, which is often unattainable in real-world settings.
While some previous works have managed to synthesize novel views given images with different illumination, their performance still relies on a substantial number of input multi-view images.
To address this problem, we suggest ExtremeNeRF, which utilizes multi-view albedo consistency, supported by geometric alignment.
Specifically, we extract intrinsic image components that should be illumination-invariant across different views, enabling direct appearance comparison between the input and novel view under unconstrained illumination. 
We offer thorough experimental results for task evaluation, employing the newly created NeRF Extreme benchmark—the first in-the-wild benchmark for novel view synthesis under multiple viewing directions and varying illuminations.

\end{abstract}
    
\section{Introduction}
Neural radiance fields (NeRF)~\cite{mildenhall2020nerf} have recently made a substantial impact on 3D vision.
Through optimizing a multi-layered perceptron (MLP) for mapping 3D point locations to color and volume density, NeRF significantly outperforms prior works~\cite{lombardi2019neural, sitzmann2019scene, mildenhall2019local} in novel view synthesis.

However, what if \textit{only a few images collected from the internet or mobile phones taken under unconstrained illumination conditions are available?} 
In most cases, NeRF-based novel view synthesis under such a practical environment is often limited since it 1) requires a massive amount of data for reliable synthesis results, and 2) assumes constrained illumination conditions among input views to encode a view-dependent color. These are key drawbacks for practical usage of NeRF, as they disable view synthesis on images that were casually collected or captured in daily life.

\begin{figure}[t]
    \centering
     \begin{subfigure}[b]{1.0\linewidth}
         \centering
        \includegraphics[width=\linewidth]{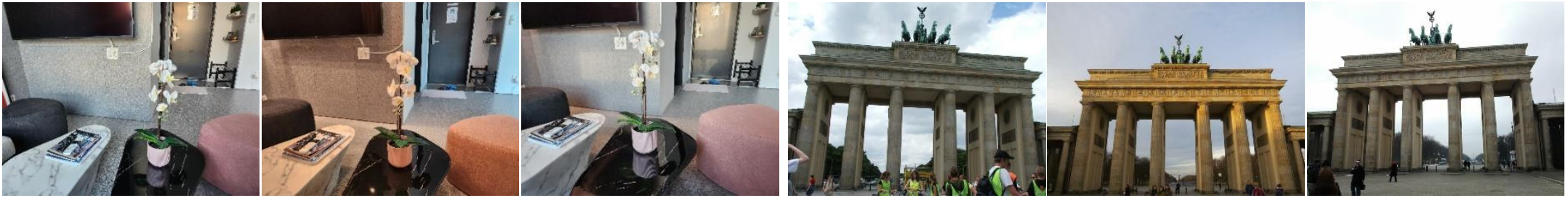}
     \end{subfigure}
    \begin{tabular}{>{\centering\arraybackslash}p{.75\linewidth}}
    \scriptsize Two sets of sparse inputs with varying illuminations \\ [0.1cm]
    \end{tabular}
     \begin{subfigure}[b]{1.0\linewidth}
         \centering
        \includegraphics[width=\linewidth]{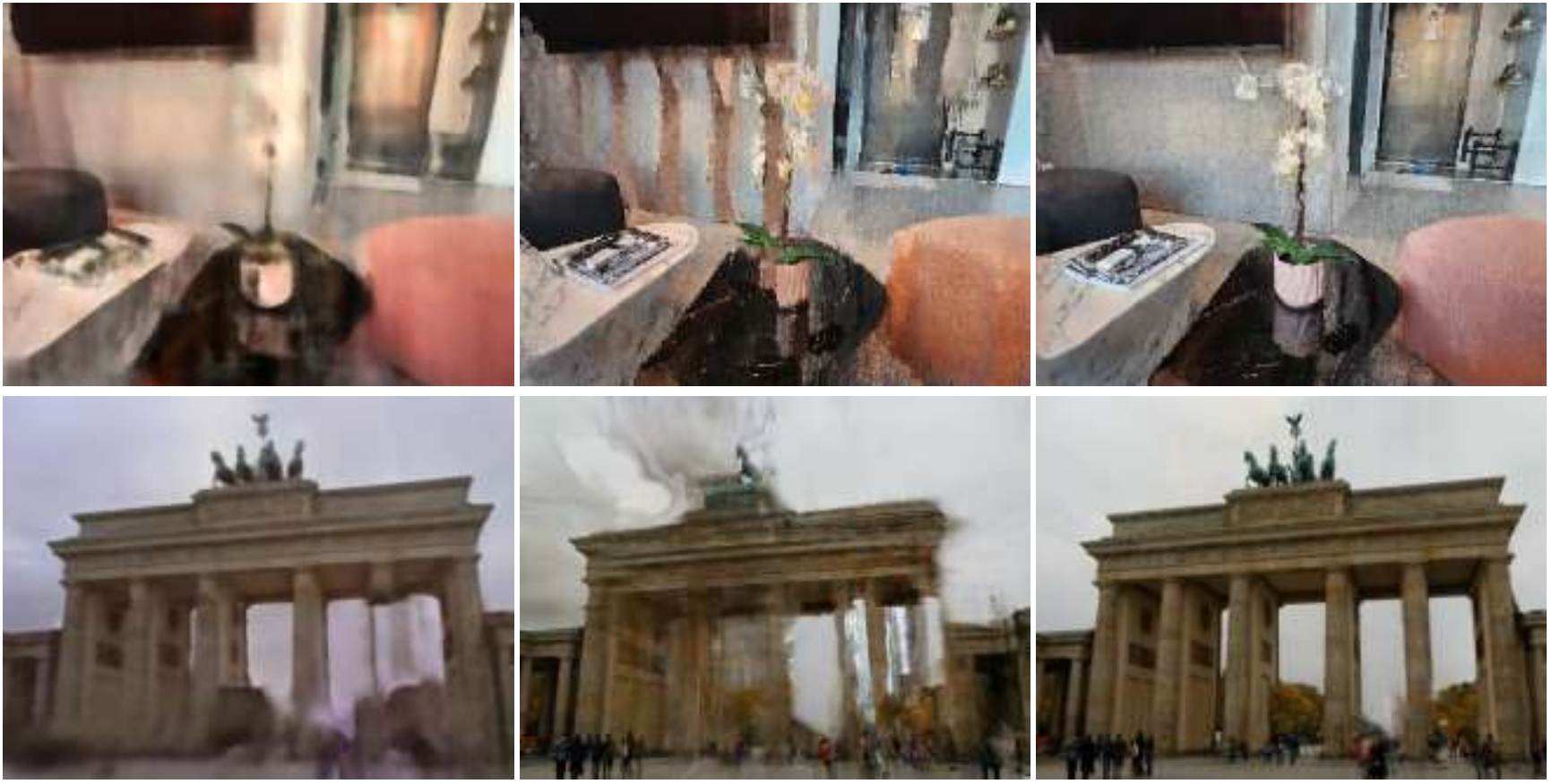}        %
     \end{subfigure}
      \begin{tabular}{>{\centering\arraybackslash}p{.28\linewidth}>{\centering\arraybackslash}p{.28\linewidth}>{\centering\arraybackslash}p{.30\linewidth}}
     \scriptsize NeRF-W~(CVPR'21) & \scriptsize RegNeRF~(CVPR'22) & \scriptsize \textbf{ExtremeNeRF (Ours)}\\
      \end{tabular}
    \caption{
        \textbf{Few-shot view synthesis results on few inputs with varying illuminations.}
    Our ExtremeNeRF demonstrates reliable results in comparison to baseline methods for two specific scenarios: NeRF under varying illuminations (NeRF-W) and few-shot view synthesis (RegNeRF).}
    \label{fig:teaser}
\end{figure}

NeRF-W~\cite{nerfinthewild} pioneered view synthesis with inputs under varying illumination, enabling novel view synthesis from internet-collected tourism images~\cite{snavely2006photo}.
Subsequent work~\cite{HaNeRF} enables appearance hallucination of the synthesized image given unconstrained image collections, by learning a view-consistent appearance of the scene. 
However, these works are hindered by the limited number of input images~(see Fig.~\ref{fig:teaser}). Moreover, previous works that deal with few-shot view synthesis~\cite{yu2021pixelnerf, dietnerf, infonerf, niemeyer2022regnerf, dsnerf, yang2023freenerf} are often hindered by the illumination variation due to the characteristic of NeRF that learns radiance dependent on viewing direction and illumination. 
.

\begin{figure}
  \centering
  \includegraphics[width=\linewidth]{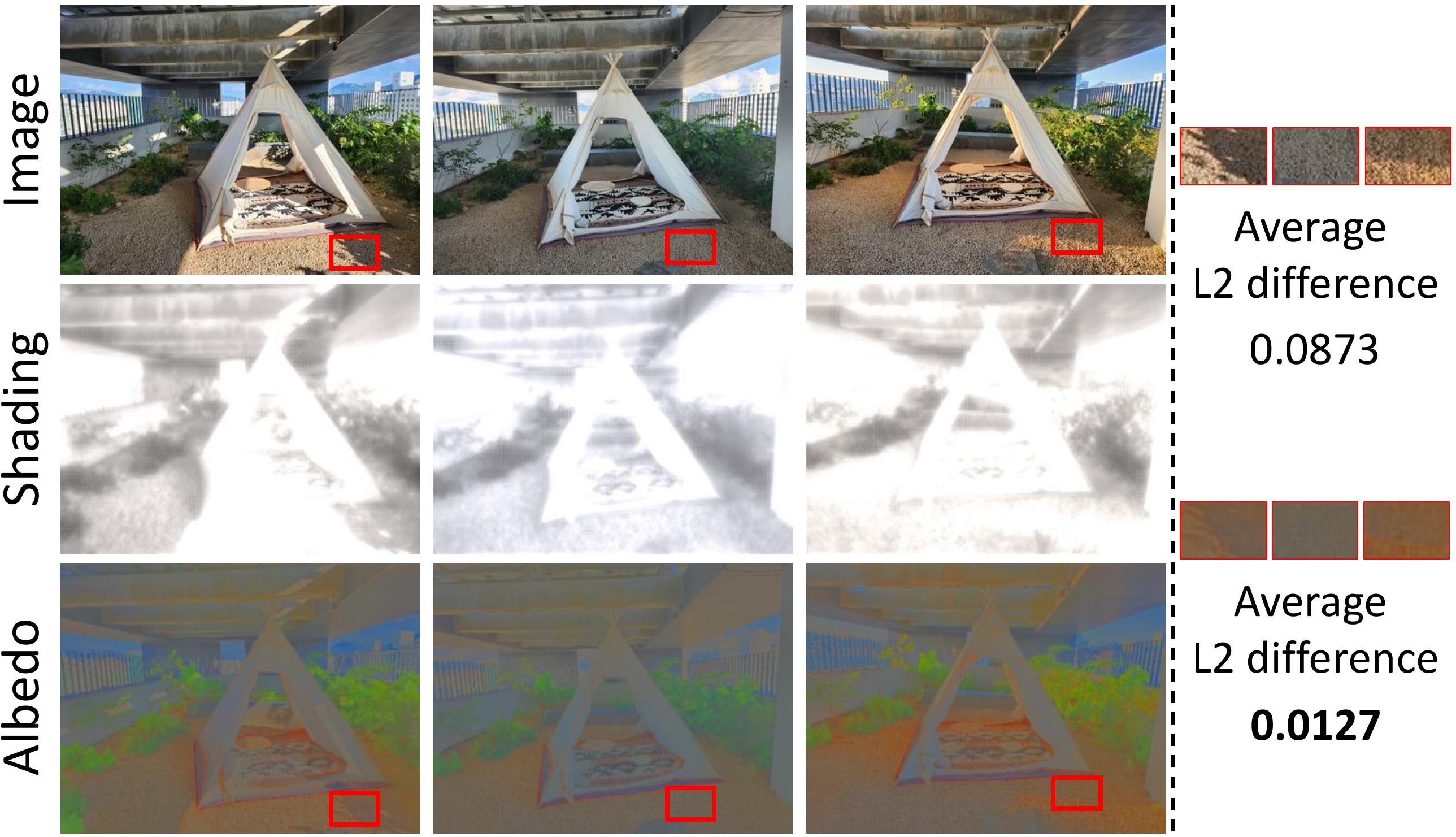}

\caption{ \textbf{Intrinsic decomposition on multi-view images under varying illumination.}
Estimated albedo maps exhibit more illumination invariance compared to color maps, resulting in lower differences across multiple views.}
\label{fig:intrinsics} 
\end{figure}

In this paper, we address the problem of novel view synthesis of scenes \textit{given only sparse input images taken under unconstrained illumination}, for the first time. 
Our proposed method, dubbed ExtremeNeRF, leverages intrinsic decomposition to mitigate the problem.
The color of the scene referred to as albedo, plays an essential role in maintaining consistency regardless of changes in viewing direction or illumination conditions~(see Fig.~\ref{fig:intrinsics}).

Since NeRF often struggles in rendering a large-size patch due to the complexity, it is challenging to infer intrinsic components from the rendered images that are largely dependent on global contexts~\cite{ye2022intrinsicnerf}. To overcome this, we first extract the global context-aware pseudo-albedo ground truth of the inputs in the offline process. By enforcing a patch-wise module to decompose the same albedo as the pseudo-ground-truth, we then achieve global context-aware intrinsic decomposition during NeRF's optimization with minimum computational costs in an end-to-end manner.
This albedo consistency loss is supported by the geometric alignment and depth consistency loss, which provides correspondences between pixels to compare and encourages correct geometry synthesis.

In evaluating our proposed method, we utilize the benchmark datasets~\cite{snavely2006photo, HaNeRF} as well as our newly developed NeRF Extreme benchmark. NeRF Extreme represents the first-of-its-kind benchmark for in-the-wild novel view synthesis, capturing scenes under multiple viewing directions and varying illumination.

\section{Related Work}
\paragraph{Neural radiance fields.}

Since the introduction of NeRF~\cite{mildenhall2020nerf}, various extensions have been proposed~\cite{Dnerf, park2021nerfies, dreamfield, poole2022dreamfusion, nerfediting, kuang2023palettenerf}. However, NeRF still relies on a massive amount of images taken under consistent illumination. 
Some of the works investigate ways to synthesize novel views with sparse input views. Yu et al.~\cite{yu2021pixelnerf} has proved that leveraging knowledge priors leads to better few-shot view synthesis. The following works~\cite{wang2021ibrnet, dietnerf, CLIPnerf, infonerf, deng2023nerdi} have suggested a variety of priors to improve the performance. 
Other works have focused on building geometry constraints to address the distortions that arise from sparse input views. Deng et al.~\cite{dsnerf} and Xu et al.~\cite{xu2022sinnerf} have presented depth prior-based methods, as~\cite{attal2021torf, roessle2022dense, geonerf, rgbdnerf}. Some of the other methods~\cite{chen2021mvsnerf, geonerf, watson2022novel, wynn2023diffusionerf} utilize implicit geometry priors for the task. Recently, RegNeRF~\cite{niemeyer2022regnerf} suggest depth smoothness constraints enhance the rendered novel view geometry, while FreeNeRF~\cite{yang2023freenerf} add frequency regularization on it. 

View synthesis with inputs taken under varying illuminations is covered by some of the previous works~\cite{nerfinthewild, HaNeRF}, however, they rely on a massive amount of input images~(see Tab.~\ref{tab:comparison}) rather than sparse input views. 

\paragraph{Illumination decomposition.}  
Various frameworks have been developed to tackle the problem of decomposing multiple scene properties including illumination, some of which rely on large datasets of paired images and ground truth information~\cite{li2020inverse, li2021openrooms, choi2023mair}. Other approaches, such as those proposed in works like~\cite{li2018learning, liu2020unsupervised, das2022pie}, have explored methods to address the problem of decomposing illumination-invariant color from the scene.
With the help of NeRF, some of the recent works~\cite{boss2021nerd, NeuralPIL, boss2022-samurai, relightNeRF} include NeROIC~\cite{kuang2022neroic} deal with a neural decomposition of an image. However, they require massive multi-view sampling of an object, rather than a scene. Decomposing illumination from a scene involves the complex interaction of indirect illumination and scene geometries, aspects that are not extensively addressed in object-level neural decomposition. Other recent works~\cite{ye2022intrinsicnerf, nerfOSR, kuang2023palettenerf, Yang_2023_CVPR} deal with NeRF-based inverse rendering of a scene, however, focusing on disentanglement of a scene component rather than view synthesis.

\begin{table}[t]
    \resizebox{\linewidth}{!}
    {\begin{tabular}{lcccccc}

\toprule
& NeRF & RegNeRF & NeRF-W & Ha-NeRF & NeROIC  & {\textbf{Ours}} \\
\midrule
Varying.-illum. & \xmark & \xmark & \cmark &\cmark & \cmark & \cmark \\
Few-shot & \xmark & \cmark & \xmark & \xmark & \cmark & \cmark \\
Non-object-centric & \cmark & \cmark & \cmark & \cmark & \xmark & \cmark\\
\bottomrule
\end{tabular}  }
    \caption{
    \textbf{Comparisons with the baselines.}
    Our proposed method enables few-shot view synthesis given in-the-wild, non-object-centric images taken under varying illumination.}
    \label{tab:comparison} 
\end{table}

\begin{figure*}[t]
  \centering
  \includegraphics[width=\linewidth]{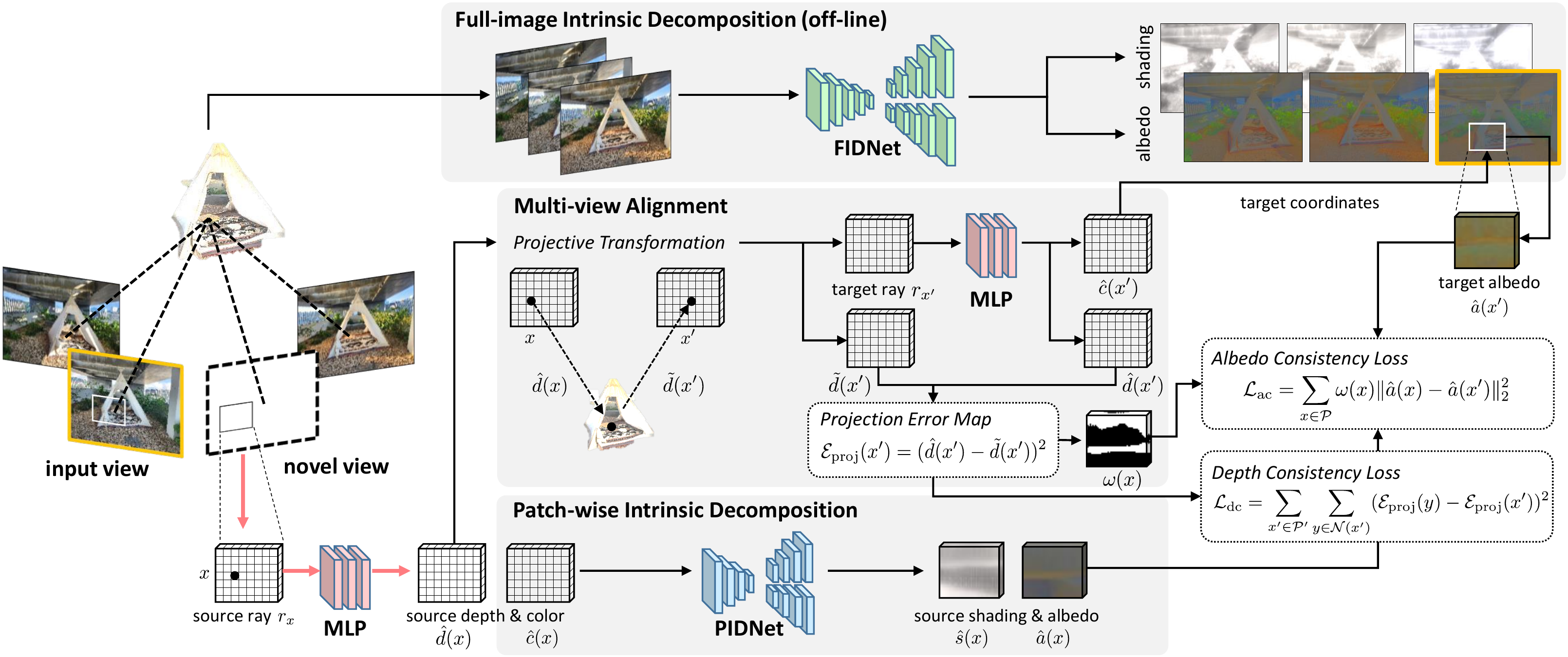}
   \caption{\textbf{Overall architecture of our ExtremeNeRF.} PIDNet extracts intrinsic components from the synthesized patch $\hat{c}(x)$ while enforcing extracted albedo to be identical with the pseudo-albedo ground truth. A weight term $\omega(x)$ and depth consistency loss $\mathcal{L}_{\mathrm{dc}}$ encourage proper correspondence matching between two views.  A bold, crimson arrow indicates the inference phase.}
   \label{fig:overallframework} 
\end{figure*}

\section{Preliminaries}
\paragraph{Neural radiance field.}
NeRF~\cite{mildenhall2020nerf} is a view-synthesis framework that maps 5D inputs (3D coordinate and viewing direction of a ray) to color and volume density, denoted by $c$ and $\sigma$, respectively.
Specifically, with a ray $r_x(t) = o + td_x$, where $o$, $d_x$, and $t$ indicate camera origin, ray direction, and scene bound at pixel location $x \in \mathbb{R}^2$, respectively, a view-dependent color $\hat{c}(x) \in {[0,1]}^3$ can be rendered such that
\begin{equation}
  \hat{c}(x) =\int_{t_n}^{t_f}T(t)\sigma(t)c(t)dt ,
  \label{eq:nerf}
\end{equation}
while $T(t) = \exp(-\int_{t_n}^{t}\sigma(s)ds)$ and $\sigma(\cdot),c(\cdot)$ are density and color predictions from the network, respectively. Similarly, a depth value $\hat{d}(x)$ at $x$ can also be rendered as
\begin{equation}
  \hat{d}(x) =\int_{t_n}^{t_f}T(t)\sigma(t)tdt .
  \label{eq:nerfdepth}
\end{equation}

Optimization in NeRF relies on a mean squared error on synthesized color $\hat{c}(x)$ as
\begin{equation}
  \mathcal{L}_{\mathrm{color}} = \sum_{x \in \mathcal{S }} \| \hat{c}(x) - c_\mathrm{gt}(x)\|^2_2 ,
  \label{eq:colorreg} 
\end{equation}
where $\mathcal{S}$ indicates the set of sampled pixels, and $c_\mathrm{gt}(x)$ indicates ground-truth color at $x$. Since volume density $\sigma$ is also optimized based on the color consistency across different views, violation of the consistent illumination assumption results in inaccurate geometry.

\paragraph{Intrinsic decomposition.}
Intrinsic decomposition aims to decompose an image into illumination-invariant color, referred to as albedo, and shading, based on the Lambertian assumption that every observed surface is diffuse. Specifically, a pixel color $c(x)$ is formulated as a multiplication of the albedo~($a(x)$) and the shading~($s(x)$) as follows:
\begin{equation}
\begin{split}
    \log c(x) = \log a(x) + \log s(x) .
\end{split}
\label{eq:imageformation1}
\end{equation}
However, most real-world objects have surfaces whose reflectances vary upon viewing directions and are often lit by colored lights. Thus, Eq.~\ref{eq:imageformation1} can be rewritten as follows:
\begin{equation}
\begin{split}
    \log c(x) = \log a(x) + \log s(x) + l + R,
\end{split}
\label{eq:imageformation2}
\end{equation}
which takes light color vector $l$ and non-Lambertian residuals $R$ into account~\cite{li2018learning}.

Real-world image intrinsic decomposition remains a challenging, imperfectly solved task. While recent works~\cite{li2018learning, liu2020unsupervised, das2022pie} demonstrate reliable performance, they still face limitations with unseen and challenging cases. Additionally, relying on global context for large-resolution image rendering in NeRF incurs computational and memory expenses.


\section{Method}

\subsection{Overview}
The objective of this work is to build an illumination-robust few-shot view synthesis framework by regularizing albedo that should be identical across multi-view images regardless of illumination. Our major challenges are to 1) achieve reliable geometry alignment between different views and 2) decompose the albedo of a rendered view without extensive computational costs.

Instead of directly addressing NeRF-based intrinsic decomposition, we integrate a pre-existing intrinsic decomposition network with NeRF optimization. Our approach involves a few-shot view synthesis framework that employs an offline intrinsic decomposition network, offering global context-aware pseudo-albedo ground truth without the computational overhead.
As illustrated in Fig.~\ref{fig:overallframework}, FIDNet provides pseudo-albedo ground truths for the input images before the start of the training, guiding PIDNet to extract intrinsic components for novel synthesized views based on these pseudo truths and multi-view correspondences. This allows our NeRF to learn illumination-robust few-shot view synthesis through cross-view albedo consistency. Subsequent subsections detail each framework component.

\subsection{Albedo Estimation}
Building upon our hypothesis that albedo aids in view synthesis with varying illumination inputs, it is crucial to decompose intrinsics from both the input and the novel views.

Instead of relying on optimization-based methods~\cite{boss2021nerd, NeuralPIL, boss2022-samurai, ye2022intrinsicnerf}, which may yield sub-optimal outcomes with limited data (0.06 times less), we propose a concise two-stage intrinsic decomposition pipeline: a full-image and patch-wise intrinsic decomposition network, called FIDNet and PIDNet, respectively. FIDNet, formulated with a pre-trained intrinsic decomposition model, extracts the albedo of the input images - pseudo-albedo ground truths - offline, to guide PIDNet with global contexts. 
Given the guidance, PIDNet extracts albedo~($\hat{a}(x)$) of the synthesized color patch with the size of $S_{\mathrm{patch}}$ at the novel view ($\hat{c}(x)$), minimizing the difference with the pseudo-ground truth, $\mathcal{L}_{\mathrm{ac}}$ (Eq.~\ref{eq:ac}), supported by multi-view alignment process described below.

\subsection{Geometry Alignment and Regularization}

For any 3D point $x^w \in \mathbb{R}^3$ in a world coordinate, a camera projection from $x^w$ to pixel location $x$ can be defined by the inverse of camera-to-world transformation $T \in SE(3)$ and camera intrinsics $K \in \mathbb{R}^{3\times3}$. Likewise, a mapping from pixel location to 3D point can be defined by the inverse operation and $d(x)$, the depth at the pixel location, as:
\begin{equation}
     x = KT^{-1}x^w, \quad  x^w = Td(x)K^{-1}\Bar{x} .
\label{eq:homography}
\end{equation}
Note that $\Bar{x} = [x^T, 1]$, a homogeneous representation of $x$.

Given a pixel $x$ in the novel view, we need the pixel $x'$ in the input view depicting the same 3D point $x^w$ as $x$ for cross-view consistency. If the depth of a given image pixel $d(x)$ is known, $x'$ can be obtained by Eq.~\ref{eq:homography} as follows:
\begin{equation}
     x' = (K'T'^{-1}T)d(x)K^{-1}\Bar{x} , 
\label{eq:homography2}
\end{equation}
where $K',T'$ and $K,T$ indicate camera intrinsics and camera-to-world matrices of the input and novel view, respectively. 

\begin{table}
    \resizebox{\linewidth}{!}{\begin{tabular}{lccccccc}
\toprule
& LLFF & DTU & PT  & NeRD & SAMURAI & ReNe & \textbf{Ours} \\
\midrule
Mult.-illum. & \xmark & \cmark & \cmark  & \cmark & \cmark & \cmark & \cmark\\
Indoor & \cmark & \cmark & \xmark  &  \cmark & \cmark & \cmark & \cmark\\
Outdoor & \cmark & \xmark & \cmark  & \cmark & \cmark & \xmark & \cmark\\
Real-world & \cmark & \cmark & \cmark  & \cmark & \cmark & \cmark & \cmark\\
In-the-wild & \cmark & \xmark & \cmark & \xmark & \cmark & \xmark & \cmark\\
Non-object-centric & \cmark & \xmark & \cmark & \xmark & \xmark & \xmark & \cmark\\
\bottomrule
\end{tabular}  
}
    \caption{
    \textbf{Multi-view dataset comparison.}
    Our NeRF Extreme dataset provides in-the-wild, non-object-centric, and varying illumination images taken indoors and outdoors.}
    \label{tab:datasets} 
\end{table}

\paragraph{Albedo consistency.}
Based on the pixel correspondence obtained above, we can impose image consistency between inputs and novel views.
However, under varying illumination, Eq.~\ref{eq:colorreg} cannot regularize view-dependent color as it does under constrained illumination, for its different interactions within illumination (see Fig.~\ref{fig:intrinsics}).
To overcome this, we present $L$-2 normalized albedo consistency loss $\mathcal{L}_{\mathrm{ac}}$ formulated as follows:
\begin{equation}
  \mathcal{L}_{\mathrm{ac}} = \sum_{x \in \mathcal{P}} \omega(x)\|\hat{a}(x) - \hat{a}(x')\|^2_2,
  \label{eq:ac}
\end{equation}
where $\hat{a}(x)$, $\hat{a}(x')$ indicate the extracted albedo from the novel and the input view, respectively, while $\mathcal{P}$ denotes all the pixels in the novel view. A weight term $\omega(x)$ is described below.

\paragraph{Visibility mask.} 
The projective transformation often utilizes incorrect synthesized depth values. For all cases, a projection error on $x'$, denote by $\mathcal{E}_\mathrm{proj}$ can be defined as follow: 
\begin{equation}    
\begin{array}{cc}
     \mathcal{E}_\mathrm{proj}(x') = (\hat{d}(x') - \Tilde{d}(x'))^2 ,
\end{array}
  \label{eq:error}
\end{equation}
where  $\hat{d}(x')$ and $\Tilde{d}(x')$ indicate rendered depth and projected depth, a byproduct of Eq.~\ref{eq:homography2}, repsectively.   
A projection error $\mathcal{E}_\mathrm{proj}$ should be close to zero if there exists neither self-occlusion nor ill-synthesized floating artifacts. 

We define visibility mask to exclude invalid cases for multi-view consistency. First, we set the mask as $0$ if the projected pixel $x'$ is outside of the field of view. Secondly, we exclude the unrelated pixel pairs caused by the scene geometry~(occlusions). To distinguish the projection errors caused by the scene geometry from the ones caused by the floating artifacts, we define a weight term $\omega$ as
\begin{equation}    
\begin{array}{cc}
     \omega(x) = \mathit{r}_{\mathrm{e}}(1 - (\mathcal{E}_\mathrm{proj}(x)/\mathcal{M}_{\mathrm{proj}})), 
\end{array}
  \label{eq:weight}
\end{equation}
where $\mathit{r}_{\mathrm{e}}$ and $\mathcal{M}_{\mathrm{proj}}$ indicate the error rate coefficient and the maximum projection error, respectively. The role of $\mathit{w}$ is to control the weight of the cross-view consistency concerning the amount of projection error. As inaccurate geometry alignments are rectified during optimization while occlusions persist, $\mathit{r}_{\mathrm{e}}$ diminishes toward the end criteria, leading to a reduction in the number of pairs that are enforced to maintain cross-view consistency.
 
\paragraph{Depth consistency.}
A direct minimization of $\mathcal{E}_\mathrm{proj}$ can be counterproductive due to occlusion, by smoothing two unrelated surface depths. Instead,  we present a depth consistency loss $\mathcal{L}_{\mathrm{dc}}$ that regularizes the amount of projection error between adjacent pixels. 
Depth consistency loss $\mathcal{L}_{\mathrm{dc}}$ in the input view can be defined such that 
\begin{equation}
     \mathcal{L}_{\mathrm{dc}} = \sum_{x' \in \mathcal{P}'} \sum_{y \in \mathcal{N}(x')}(\mathcal{E}_\mathrm{proj}(y)-\mathcal{E}_\mathrm{proj}(x'))^2 ,
\label{eq:dc}
\end{equation}
where $y$ indicates one of the 4-neighbor adjacent pixels $\mathcal{N}(x')$ for $x'$. $\mathcal{P}'$ denotes all the pixels in the input view.

\begin{figure*}[hbt!]
    \centering
     \begin{tabular}{>{\centering\arraybackslash}p{0.32\textwidth}>{\centering\arraybackslash}p{0.28\textwidth}>{\centering\arraybackslash}p{0.32\textwidth}}
      \scriptsize \textbf{NeRF under varying-illumination} & \scriptsize \textbf{Few-shot NeRF} & \scriptsize \textbf{Few-shot NeRF under varying-illumination}\\
    \end{tabular}
    \begin{tabular}{>{\centering\arraybackslash}p{0.13\textwidth}>{\centering\arraybackslash}p{0.15\textwidth}>{\centering\arraybackslash}p{0.15\textwidth}>{\centering\arraybackslash}p{0.14\textwidth}>
    {\centering\arraybackslash}p{0.15\textwidth}>
    {\centering\arraybackslash}p{0.13\textwidth}}
    \scriptsize NeRF-W~(CVPR'21) & \scriptsize Ha-NeRF~(CVPR'22) & \scriptsize RegNeRF~(CVPR'22) & \scriptsize FreeNeRF~(CVPR'23)  & 
    \scriptsize NeROIC-Geom.~(SIG'22) &
    \scriptsize  \textbf{ExtremeNeRF~(Ours)}\\
    \end{tabular}
     \begin{subfigure}[b]{1.0\textwidth} 
         \centering
        \includegraphics[width=\linewidth]{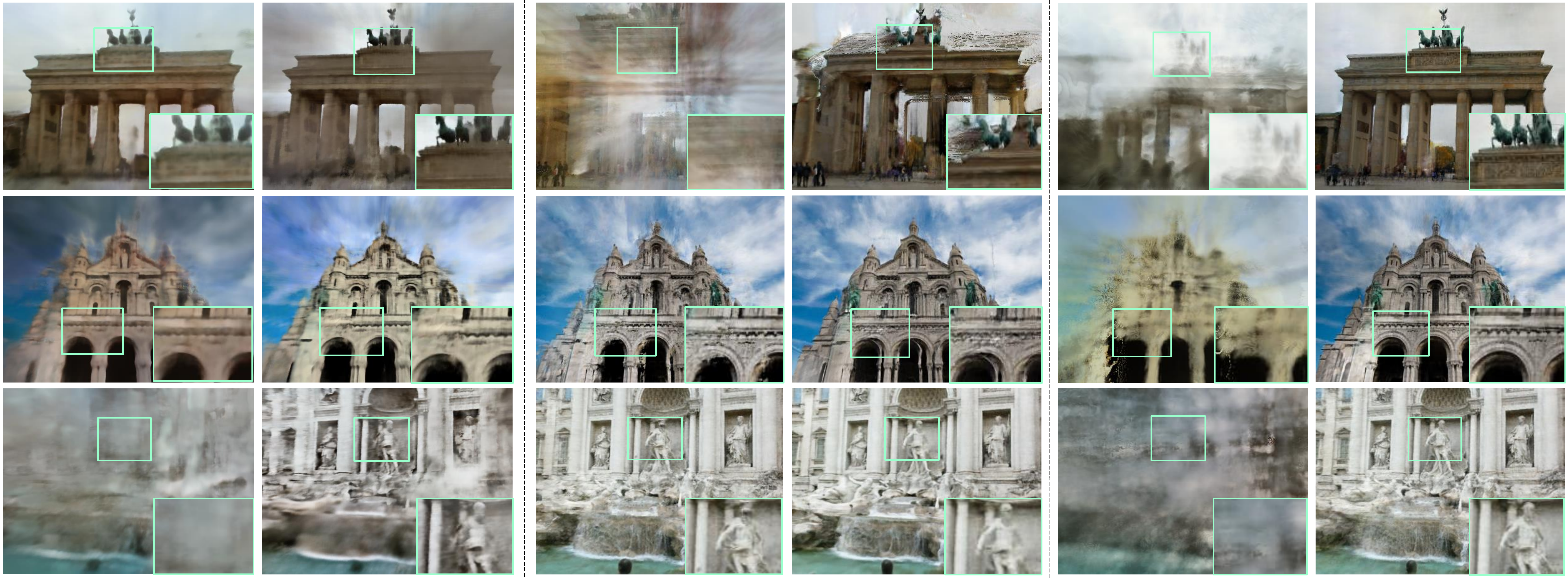}
     \end{subfigure}
    \caption{
    \textbf{Qualitative comparison on Phototourism $\mathbf{F}^3$ benchmark.}
    Synthesized novel views of `Brandenburg Gate', `Sacre Coeur', and `Trevi Fountain' (from top to bottom), generated by the baselines and our proposed method in $3$ view input images. Our method shows plausible synthesis results compared to all the baselines~(Best viewed in color). 
    }
    \label{fig:result_tourism} 
\end{figure*}

\begin{table*}[hbt!]
\centering
\resizebox{\linewidth}{!}
{\begin{tabular}{l|ccc|ccc|ccc}
\toprule
 & \multicolumn{3}{c|}{Brandenburg Gate} & \multicolumn{3}{c|}{Sacre Coeur} & \multicolumn{3}{c}{Trevi Fountain}\\ \midrule
  & SSIM $\uparrow$ & LPIPS $\downarrow$ & Abs Rel $\downarrow$ & SSIM $\uparrow$& LPIPS $\downarrow$ & Abs Rel $\downarrow$ & SSIM $\uparrow$& LPIPS $\downarrow$ & Abs Rel $\downarrow$ \\ \midrule
NeRF-W~\cite{nerfinthewild} \ & 0.39 & 0.59& \underline{0.84} & \underline{0.46} & 0.52 & \underline{0.94} & 0.16 & 0.64 &  0.65\\
Ha-NeRF~\cite{HaNeRF}\ & 0.50 & 0.43 &\textbf{0.78}  &  \underline{0.46} & 0.52 & \underline{0.94} & 0.39 & 0.48 & \textbf{0.52} \\
\midrule
RegNeRF~\cite{niemeyer2022regnerf}\ & 0.27 & 0.56& 3.54 & 0.39 & \underline{0.44} & 2.44 & 0.43 & \underline{0.37} & 0.64 \\
FreeNeRF~\cite{yang2023freenerf} \ &0.31 & 0.50 & 4.53 & 0.32 & 0.45 & 3.62 & \underline{0.45} & \textbf{0.36} & \underline{0.57} \\
\midrule
NeROIC-Geom.~\cite{kuang2022neroic}\ & 0.30 & 0.63&  0.89 & 0.34 & 0.66 & \textbf{0.85} & 0.11 & 0.70 & 0.79\\
\textbf{ExtremeNeRF (Ours)}\ & \textbf{0.56} &\textbf{0.36}& \textbf{0.78}  & \textbf{0.49} &\textbf{0.38} & 1.28 & \textbf{0.57} & \textbf{0.36} & 0.59\\
\bottomrule
\end{tabular}
}
    \caption{
    \textbf{Quantitative comparison on Phototourism $\mathbf{F}^3$ benchmark.}
   Improvements in SSIM and LPIPS prove that our model with albedo consistency succeeded in synthesizing fine geometry details.
    }
    \label{tab:tourism} 
\end{table*}

\begin{figure*}[t]
    \centering
    \begin{tabular}{>{\centering\arraybackslash}p{0.32\textwidth}>{\centering\arraybackslash}p{0.28\textwidth}>{\centering\arraybackslash}p{0.32\textwidth}}
      \scriptsize \textbf{NeRF under varying-illumination} & \scriptsize \textbf{Few-shot NeRF} & \scriptsize \textbf{Few-shot NeRF under varying-illumination}\\
    \end{tabular}
    \begin{tabular}{>{\centering\arraybackslash}p{0.13\textwidth}>{\centering\arraybackslash}p{0.15\textwidth}>{\centering\arraybackslash}p{0.15\textwidth}>{\centering\arraybackslash}p{0.14\textwidth}>
    {\centering\arraybackslash}p{0.15\textwidth}>
    {\centering\arraybackslash}p{0.13\textwidth}}
    \scriptsize NeRF-W~(CVPR'21) & \scriptsize Ha-NeRF~(CVPR'22) & \scriptsize RegNeRF~(CVPR'22) & \scriptsize FreeNeRF~(CVPR'23)  & 
    \scriptsize NeROIC-Geom.~(SIG'22) &
    \scriptsize  \textbf{ExtremeNeRF~(Ours)}\\
    \end{tabular}
     \begin{subfigure}[b]{1.0\textwidth} 
         \centering
        \includegraphics[width=\linewidth]{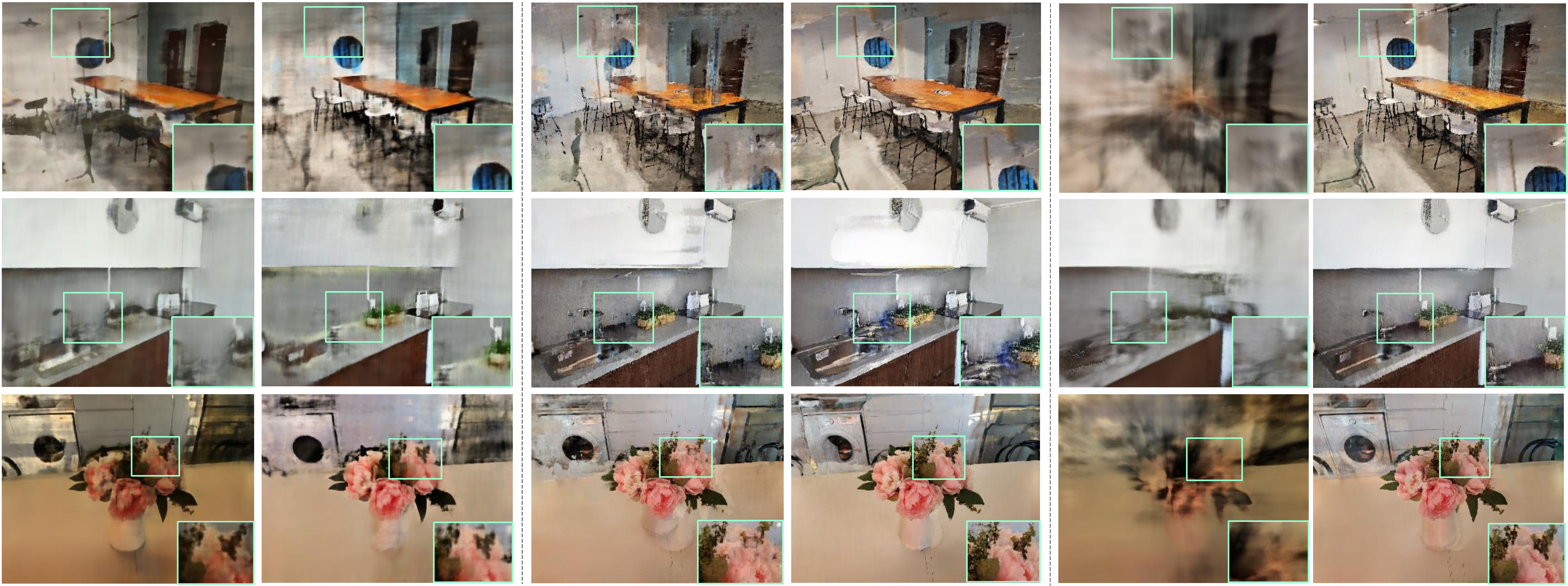}
     \end{subfigure}

    \caption{
    \textbf{Qualitative comparison on NeRF Extreme benchmark.}
    A synthesized novel view of `Cafe', `Kitchen', and `Flower'~(from top to bottom), generated by the baselines and our proposed method. Our proposed method shows strength in maintaining fine geometry details compared to the baselines~(Best viewed in color). 
    }
    \label{fig:result_extreme} 
\end{figure*}

\begin{table*}[t]
\resizebox{\linewidth}{!}{\begin{tabular}{l|ccc|ccc|ccc}
\toprule
 & \multicolumn{3}{c|}{Cafe} & \multicolumn{3}{c|}{Kitchen} & \multicolumn{3}{c}{Flower}\\ \midrule
  & SSIM $\uparrow$ & LPIPS $\downarrow$ & Abs Rel $\downarrow$ & SSIM $\uparrow$& LPIPS $\downarrow$ & Abs Rel $\downarrow$ & SSIM $\uparrow$& LPIPS $\downarrow$ & Abs Rel $\downarrow$\\ \midrule
NeRF-W~\cite{nerfinthewild} \ &0.32 & 0.55& 0.64  & \underline{0.60} & 0.47 & \underline{0.35} & 0.62& \underline{0.42} & 0.55 \\
Ha-NeRF~\cite{HaNeRF} \ & 0.36 & 0.54 & \underline{0.62} & 0.54 & 0.52 & 0.37 & \underline{0.65} & 0.43& 0.70 \\
\midrule
RegNeRF~\cite{niemeyer2022regnerf}\ & 0.36 & 0.48 & 0.66 & 0.55 & \underline{0.39} & \textbf{0.34} & 0.58 & 0.49 & 0.78 \\
FreeNeRF~\cite{yang2023freenerf} \ & \underline{0.39} & \underline{0.43} & 0.90 & 0.55 & 0.40 & 0.81 & 0.60 & \underline{0.42} & 0.86 \\
\midrule
NeROIC-Geom.~\cite{kuang2022neroic}\ & 0.25 & 0.67& 0.80  & 0.47 & 0.58 & 0.49 & 0.49 & 0.55& \underline{0.54} \\
\textbf{ExtremeNeRF (Ours)}\ &\textbf{0.48} & \textbf{0.38} & \textbf{0.51} & \textbf{0.62} & \textbf{0.34} & \underline{0.35}  & \textbf{0.67 }&\textbf{0.40}& \textbf{0.49}  \\
\bottomrule
\end{tabular}
}
    \caption{
    \textbf{Quantitative comparison on NeRF Extreme benchmark.}
    We compare our quantitative result with the baseline methods in $3$ view settings. For every case, our model shows the best synthesizing performance for each metric.
    }
    \label{tab:extreme} 
\end{table*}

\subsection{Total Loss}
In addition to the loss functions $\mathcal{L}_{\mathrm{ac}}$  and $\mathcal{L}_{\mathrm{dc}}$, we incorporate several constraints for the optimization of NeRF and PIDNet: the depth smoothness loss $\mathcal{L}_{\mathrm{ds}}$~\cite{niemeyer2022regnerf}, the edge-preserving loss $\mathcal{L}_{\mathrm{edge}}$~\cite{godard2017unsupervised}, the intrinsic smoothness loss $\mathcal{L}_{\mathrm{pid}}$~\cite{li2018learning}, the chromaticity consistency loss $\mathcal{L}_{\mathrm{chrom}}$~\cite{ye2022intrinsicnerf}, and the frequency regularization mask, proposed by FreeNeRF~\cite{yang2023freenerf}. Further details are provided in the supplementary material.


\section{Experiments}
In this section, we provide extensive comparisons with the baselines using our newly proposed datasets. Further results and details can be found in the supplementary material.
\subsection{Datasets}
In Table~\ref{tab:datasets}, Phototourism (PT) and its variants~\cite{snavely2006photo, HaNeRF} are the only benchmarks that exhibit both pose and illumination variations. However, these datasets are not suitable for few-shot view synthesis due to their randomness.
For extensive experiments, we construct two datasets for the evaluation of few-shot view synthesis under varying illumination.

\paragraph{Phototourism $\mathbf{F}^3$.}
Phototourism $\mathbf{F}^3$(Frontal Facing Few-shot), a subset of Phototourism~\cite{snavely2006photo} dataset, is specifically curated for evaluating few-shot view synthesis under varying illumination. Frontal-facing scenes within similar depth bounds and significant illumination variation in `Brandenburg Gate', `Sacre Coeur', and `Trevi Fountain' are selected for the task. The ground truth depth maps are provided by Phototourism. The rationale behind building a frontal-facing subset can be found in the supplementary material.

\paragraph{NeRF Extreme.}
To build a benchmark that fully reflects unconstrained environments, we collected multi-view images with varying light sources such as multiple light bulbs and the sun using the the mobile phone camera. 
We took $40$ images per scene - 30 images in the train set and 10 images in the test set. 
The training sets are captured with at least three different lighting conditions. 
The camera poses and depth maps are obtained using the COLMAP~\cite{colmap} and multi-view stereo method~\cite{giang2022curvatureguided}, respectively. NeRF Extreme is the first in-the-wild multi-view dataset with varying illumination, whose scenes are not limited to object-centric or outdoor scenes.

\subsection{Experimental Settings}
\paragraph{Baselines.}
Since there is no previous work that deals with scene-level few-shot view synthesis under varying illumination, we compare our proposed method against three types of baselines. 1) NeRF under varying illumination: NeRF-W~\cite{nerfinthewild}, Ha-NeRF~\cite{HaNeRF}, 2) Few-shot NeRF: RegNeRF~\cite{niemeyer2022regnerf}, FreeNeRF~\cite{yang2023freenerf}, and 3) Few-shot NeRF under varying-illumination: NeROIC~\cite{kuang2022neroic}. For NeROIC, we report NeROIC-Geom results as NeROIC-Full exhibits some divergence. Note that NeROIC is tailored for object-centric scenes, not frontal-facing ones.

For comparison, we used the mean SSIM, LPIPS metric of the synthesized image, and Abs Rel (Absolute Relative Error) of the synthesized depth map. Similar works~\cite{nerfinthewild, HaNeRF, kuang2022neroic} have evaluated performance using PSNR after relighting to match the target illumination.
However, our main aim is to highlight improved geometry details rather than relighting. Moreover, the baselines struggle with proper relighting in a few-shot setting, making PSNR unsuitable for evaluation.

\paragraph{Implementation details.}
Our framework is based on the implementation of RegNeRF~\cite{niemeyer2022regnerf}. For FIDNet, the official code and model of IIDWW~\cite{li2018learning} trained with BigTimes dataset are used without fine-tuning. An image size of $300 \times 400$ is used for the training, with~$S_{\mathrm{patch}} = 32 \times 32$. 
We train every scene for $70K$ using 4 NVIDIA A100 GPUs.

\paragraph{Computational complexity.}
Except for a few-shot NeRF, most methods require about 10 to 20 hours of training time to achieve optimal performance on 4 NVIDIA A100 GPUs. For few-shot NeRFs, RegNeRF~\cite{niemeyer2022regnerf}, our proposed method, and FreeNeRF~\cite{yang2023freenerf} take 2,4 and 1.5 hours in 3 views, respectively. 

\begin{table*}[t]
    \centering
    \resizebox{\linewidth}{!}
    {\begin{tabular}{l|ccc|ccc|ccc}
\toprule
 & \multicolumn{3}{c|}{Cafe} & \multicolumn{3}{c|}{Kitchen} & \multicolumn{3}{c}{Flower}\\ \midrule
  & SSIM $\uparrow$ & LPIPS $\downarrow$ & Abs Rel $\downarrow$ & SSIM $\uparrow$& LPIPS $\downarrow$& Abs Rel $\downarrow$ & SSIM $\uparrow$& LPIPS $\downarrow$& Abs Rel $\downarrow$ \\ \midrule 
1-1. w/o AC~(Albedo Consistency)\ & 0.414 & 0.42 & 0.92  & 0.60 & 0.35 & 0.79  & 0.63 & 0.41 & 0.86 \\
1-2. w/ AC - w/o FIDNet \ & 0.413 & 0.42 & 0.92  & 0.60 & 0.35 & 0.80 & 0.63 & 0.42 & 0.88 \\
1-3. w/ AC - Albedo MLP \ & 0.413 & 0.42 & 0.92 & 0.59 & 0.35 & 0.85  & 0.64 & 0.41 & 0.89 \\
1-4. w/ AC~\cite{das2022pie} \ & 0.445 & 0.39 & 0.89 & 0.60 & 0.35 & 0.79 & 0.62 & 0.42 & 0.88 \\
\midrule
2-1. w/o DC~(Depth Consistency)\ & 0.470 & 0.39 & 0.89 & \textbf{0.62} & 0.35 & 0.81  & 0.62 & 0.44 & 0.88 \\
2-2. w/o Visibility Mask \ & 0.404 & 0.43 & 0.92  & 0.60 & 0.36 & 0.80 & 0.62 & 0.43 & 0.89  \\
\midrule
\textbf{ExtremeNeRF (Ours)} \ & \textbf{0.476} & \textbf{0.38} & 0.51 & \textbf{0.62} & \textbf{0.34} & \textbf{0.36} & \textbf{0.67} & \textbf{0.40} & \textbf{0.49} \\
\bottomrule
\end{tabular}
}
    \caption{
    \textbf{Ablation study of our ExtremeNeRF.}
   Ablation studies on two different groups are provided, in terms of 1) albedo consistency and 2) depth consistency.}
    \label{tab:ablation} 
\end{table*}

\subsection{Experimental Results}
\begin{figure}
    \centering
    \begin{tabular}{>{\centering\arraybackslash}p{0.15\linewidth}>{\centering\arraybackslash}p{0.21\linewidth}>{\centering\arraybackslash}p{0.17\linewidth}>{\centering\arraybackslash}p{0.20\linewidth}}
      \scriptsize Ha-NeRF & \scriptsize  FreeNeRF & \scriptsize  Ours w/o AC & \scriptsize  \textbf{ExtremeNeRF~(Ours)}\\
    \end{tabular}
     \begin{subfigure}[b]{1.0\linewidth} 
         \centering
        \includegraphics[width=\linewidth]{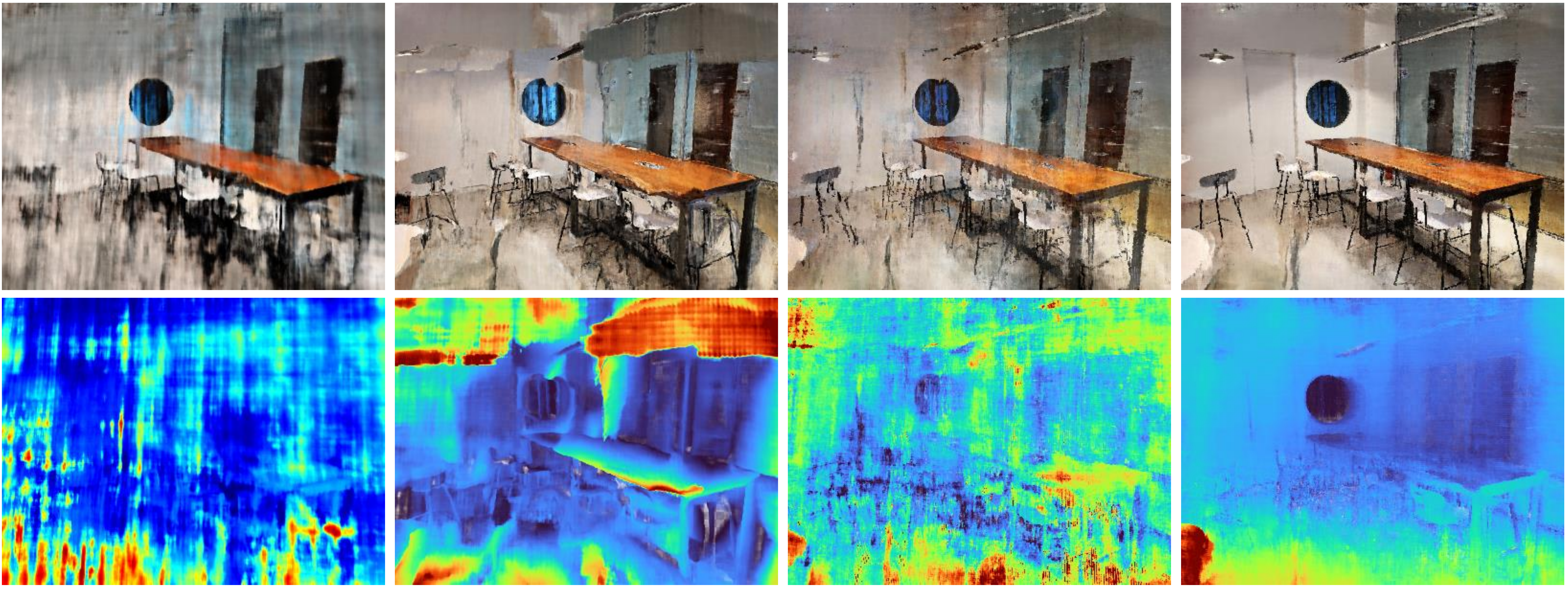}
     \end{subfigure}
    \caption{
    \textbf{Depth map comparison.}
   Depth maps paired with synthesized images of `Cafe' scene of NeRF Extreme benchmark are selected~(Best viewed in color). 
    }
    \label{fig:depthmap} 
\end{figure}

\paragraph{Comparisons with the baselines.}
Fig.~\ref{fig:result_tourism} and Fig.~\ref{fig:result_extreme} show the qualitative comparison between our ExtremeNeRF and other baseline methods on Phototourism $\mathrm{F}^3$ and NeRF Extreme, respectively. In the $1$st and $2$nd rows, baselines dealing with varying illumination lack input images, resulting in smoothed geometry details.
For few-shot NeRF methods (the $3$rd and $4$th rows), synthesized geometries face challenges due to inconsistent illumination. Particularly, baselines exhibit higher distortion when confronted with significant illumination variations, as observed in the `Brandenburg Gate' and `Cafe' scenes, respectively. 
The results are further supported by quantitative comparisons in Tab.\ref{tab:tourism} and Tab.\ref{tab:extreme}, especially with a large improvement in SSIM and LPIPS~(bold texts for the best performance, and underline for the 2nd best). 
In the case of NeROIC~\cite{kuang2022neroic}, synthesized results from NeROIC-Geom., which is a partially optimized version of the method, are reported. Note that the entire model shows diverged results on frontal-facing scenes. 
In all cases, our method demonstrates plausible synthesized results with fine geometry details. Additionally, Fig.~\ref{fig:depthmap} illustrates that our model exhibits reliable depth synthesis, leading to the expectation of achieving plausible video synthesis performance, even when the Abs Rel score is compatible with each other~(`Brandenburg Gate' scene).

\begin{table}
    \centering
    \resizebox{\linewidth}{!}
    {\begin{tabular}{l|ccc}
\toprule
  & SSIM $\uparrow$ & LPIPS $\downarrow$ & Abs Rel $\downarrow$  \\ \midrule
NeRF-W~\cite{nerfinthewild} \ & 0.38 & 0.51 & 0.79\\
RegNeRF~\cite{niemeyer2022regnerf} \ & 0.44 & 0.35 & \bf{0.76} \\
\textbf{ExtremeNeRF~(Ours)} \ & \bf{0.45} & \bf{0.34} & \bf{0.76} \\
\bottomrule
\end{tabular}
}
    \caption{
    \textbf{Quantitative Comparison on LLFF.} Quantitative comparison on the `fern' scene of the LLFF~\cite{mildenhall2020nerf} in 3 view settings. }
    \label{tab:llff} 
\end{table}

\paragraph{Ablation studies.}
Tab.~\ref{tab:ablation} shows groups of ablation studies to validate the design choices of our work. Each studies related to albedo consistency~(1-1 to 1-4) and depth consistency~(2-1 to 2-2). The additional ablation studies on the patch size and learned priors can be found in the supplementary material.

The experiments in the first group demonstrate that incorporating albedo consistency between the input and novel views contributes to the regularization of geometry. 
Quantitative results in 1-2, reveal that removing albedo consistency leads to sub-optimal performance, indicating the importance of this constraint for well-constrained optimization of a novel view. A qualitative comparison of the synthesized maps in Fig.~\ref{fig:depthmap} supports the idea that incorporating albedo consistency contributes to reliable depth estimation. 
In 1-3 and 1-4, we provide empirical evidence that FIDNet serves as a suitable guide for achieving cross-view consistency, rather than MLP that synthesizes albedo. In 1-4, we replace the FIDNet model from IIDWW~\cite{li2018learning} with the other intrinsic decomposition model~\cite{das2022pie}, however, shows degraded performance. 
Note that FIDNet can be substituted with other models in our framework if they exhibit superior performance.

The experiments in the second group illustrate that depth consistency, when taken into account with proper consideration of scene geometry, contributes to improved geometry. Ablating depth consistency (2-1) and visibility mask (2-2) results in unreliable depths, as enforcing consistency between unrelated surfaces leads to undesirable outcomes. 

\paragraph{Comparisons on LLFF.}
To assess performance on the benchmark with constrained illumination, Table~\ref{tab:llff} compares results on the 'fern' scene from the LLFF~\cite{mildenhall2019local} dataset. RegNeRF~\cite{niemeyer2022regnerf} shows minor differences compared to our method when illumination is shared among inputs, while NeRF-W~\cite{nerfinthewild} significantly degrades with few-shot inputs.

\section{Conclusion and Further Work}
In this paper, we proposed ExtremeNeRF, which can synthesize a novel view in practical environments, where neither a large amount of multi-view images nor consistent illumination is available. By regularizing albedo which should be identical across different views, our method can directly regularize appearance instead of interpolating view-dependent color as vanilla-NeRF did. We have proved that the proposed method outperforms other previous works with new benchmarks in a few-shot view synthesis under an unconstrained illumination environment. Any few-shot NeRF can obtain illumination-robust regularization by utilizing our proposed albedo consistency constraints on their optimization. However, similar to other optimization-based NeRF approaches, relighting a scene given sparse inputs remains a challenge. Further, significant illumination variation may result in noisy input camera poses. Addressing these problems could be a potential direction for our future work.

\section{Acknowledgments}
This work was supported by Institute of Information \& communications Technology Planning \& Evaluation (IITP) grant funded by the Korea government(MSIT) (No.RS-2023-00227592, Development of 3D object identification technology robust to viewpoint changes, 70\%, and No.2020-0-00457, Development of free-form plenoptic video authoring and visualization platform for large space, 30\%)

\bibliography{aaai24}

\clearpage


\newpage
\clearpage

\maketitle
\appendix
\section{Supplementary Material}

In this section, we provide further details on the experiments and datasets, followed by additional ablation studies and experimental results.

\section{More Details on Experiments}
\subsection{Experimental Details}
\begin{figure}[t]
    \centering
         \centering
        \includegraphics[width=\linewidth]{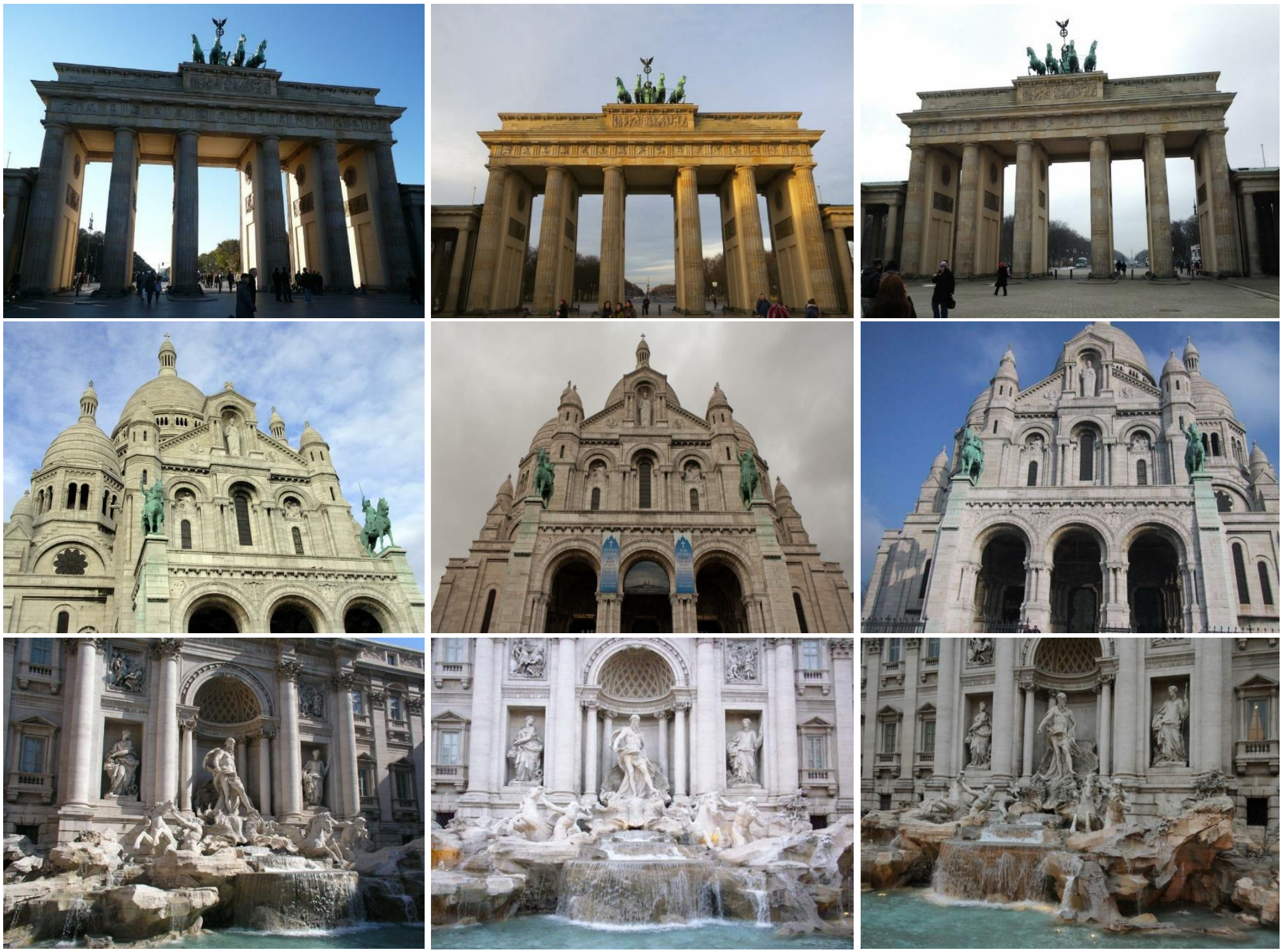}
    \caption{
    \textbf{Examples of inputs sampled from Phototourism $\mathbf{F}^3$.}
   Sampled frontal-facing scenes with varying illumination from the `Brandenburg Gate',   `Sacre Coeur', and `Trevi Fountain', respectively. 
    }
    \label{fig:inputs_tourism} 
\end{figure}

\paragraph{Comparisons on Phototourism $\mathbf{F}^3$.}
A subset for few-shot view synthesis was created by selecting 3 input views with similar depth bounds and frontal-facing poses. The image IDs are (185, 45, 1066), (964, 34, 478), and (82, 312, 803) for `Brandenburg Gate', `Sacre Coeur', and `Trevi Fountain', respectively. Fig.~\ref{fig:inputs_tourism} shows image samples of the dataset. Each scene has about eight test images to evaluate the performance. For the depth map comparison using Abs Rel, the original ground truth depth maps provided by Phototourism~\cite{snavely2006photo} are very noisy. Following the publicly available instructions of the dataset, we used the clean versions of the depth maps with the background masks. As a result, reported Abs Rel excludes background regions for the evaluation. More detailed information about the proposed datasets can be found in the next section.

\paragraph{Comparisons on NeRF Extreme.}
For evaluating the few-shot view synthesis performance on NeRF Extreme, image IDs (0, 14, 29) were used as inputs for each scene. Given that our proposed multi-view consistency takes into account complex scene geometry, including occlusions, the optimal model and parameters might vary based on the specific scene characteristics. However, experimental results in Tab.~$4$ of the paper were based on our final model.

\paragraph{Ablation studies.}
In this paragraph, we provide detailed information about the ablation studies, especially for ablation 1-3, with albedo MLP. Ablations other than 1-3, were conducted with straightforward implementation with and without proposed consistency regularization. In the case of 1-3, we implemented the multi-view albedo consistency by directly synthesizing the albedo map using the Multi-Layered Perceptron (MLP) of NeRF. Given the view-independent nature of albedo, we configured the MLP to output albedo as density. By doing so, the proposed framework may be free from the computational costs that come from continuous patch-wise sampling. However, as shown in Tab.~5 in the paper, the model with albedo MLP shows sub-optimal results, which has almost the same results as the model without albedo consistency~(1-1).

\subsection{Implementation Details}
\paragraph{Neural radiance fields.}
Our framework is based on JAX~\cite{jax2018github} implementation of RegNeRF~\cite{niemeyer2022regnerf}, while partially adopting a frequency regularization mask of FreeNeRF~\cite{yang2023freenerf} when constructing MLP. Detailed algorithms of the proposed loss functions are provided in Alg.~\ref{code:dcloss} and \ref{code:acloss}.

\paragraph{Intrinsic decomposition network.}
Building upon the concept of integrating an offline intrinsic decomposition network, our PIDNet adopts the architecture of the chosen FIDNet. Given that our ultimate model employs IIDWW~\cite{li2018learning} as the FIDNet to provide pseudo-albedo ground truth, our PIDNet shares a similar architecture with IIDWW, albeit in a shallower configuration.
However, it's worth highlighting that any intrinsic decomposition network demonstrating superior performance can substitute IIDWW as the FIDNet with a paired PIDNet that has a similar, and shallower architecture.

\begin{algorithm}[hbt]
\caption{Depth Consistency, Pytorch-like}
\label{code:dcloss}
\definecolor{codeblue}{rgb}{0.25,0.5,0.5}
\definecolor{codekw}{rgb}{0.85, 0.18, 0.50}

\lstset{
  backgroundcolor=\color{white},
  basicstyle=\fontsize{7.5pt}{7.5pt}\ttfamily\selectfont,
  columns=fullflexible,
  breaklines=true,
  captionpos=b,
  commentstyle=\fontsize{7.5pt}{7.5pt}\color{codeblue},
  keywordstyle=\fontsize{7.5pt}{7.5pt}\color{codekw},
}
\begin{lstlisting}[language=python]
def get_corresponding_coords(K, coords, c2w_1, c2w_2, depth):
    im_coord = coords*depth
    cam_coord = np.linalg.inv(K) @ im_coord
    w2c = np.linalg.inv(c2w_2)
    cam_mat = w2c @ c2w_1
    return K @ cam_mat @ cam_coord

def init_mask_tg_rays(tg_coords, full_rays):
    mask = np.ones_like(tg_coords)
    mask = (0 if outside of image region else 1)
    tg_rays = full_rays[tg_coords]
    return lrc_mask, tg_rays

# src renderings from MLP
src_renderings = model.apply(src_rays)
depth = src_renderings['depth_pred']

# calc target coord & rays
uvd_target = get_corresponding_coords
          (K, src_coords, src_pose, tg_pose, depth)
tg_coords = uvd_target[:2,:]//uvd_target[2,:]

depth_tilde = uvd_target[2,:]
mask, tg_rays = init_mask_tg_rays
          (tg_coords, tg_full_rays)

# tg renderings from MLP
tg_renderings = model.apply(tg_rays)
depth_hat = tg_renderings['depth_pred']

error_proj = (depth_tilde - depth_hat)**2

loss_dc = np.mean(tv_norm(error_proj))
\end{lstlisting}
\end{algorithm}

\begin{algorithm}[hbt]
\caption{Albedo Consistency, Pytorch-like}
\label{code:acloss}
\definecolor{codeblue}{rgb}{0.25,0.5,0.5}
\definecolor{codekw}{rgb}{0.85, 0.18, 0.50}

\lstset{
  backgroundcolor=\color{white},
  basicstyle=\fontsize{7.5pt}{7.5pt}\ttfamily\selectfont,
  columns=fullflexible,
  breaklines=true,
  captionpos=b,
  commentstyle=\fontsize{7.5pt}{7.5pt}\color{codeblue},
  keywordstyle=\fontsize{7.5pt}{7.5pt}\color{codekw},
}
\begin{lstlisting}[language=python]

src_patch = src_renderings['rgb']
src_shading, src_albedo, src_rgb = pidnet.apply(src_patch)

# intrinsic smoothness loss
pid_loss = np.mean(tv_norm(src_albedo))

# get target albedo from FIDNet result
tg_albedo = tg_full_albedo[tg_coords]

tg_chrome = rgb_to_chromaticity(tg_albedo)
src_chrome = rgb_to_chromaticity(src_albedo)
src_patch_chrome = rgb_to_chromaticity(src_patch)

# chromaticity consistency loss
loss_chrom = np.mean((src_chrome - tg_chrome)**2 
            + (src_chrome - src_patch_chrome)**2)

# scale value using least square
tg_albedo = ls_scale_val(src_albedo, tg_albedo)

occ_weight = r_e * (1- error_proj/max(error_proj))

# albedo consistency loss
loss_ac = np.mean(occ_weight*(src_albedo - tg_albedo)**2)

src_grad = np.gradient(np.exp(mask*src_albedo))
tg_grad = np.gradient(np.exp(mask*tg_albedo))

# edge preserving loss
loss_edge = np.mean(occ_weight*(src_grad - tg_grad)**2)

\end{lstlisting}
\end{algorithm}

\subsection{Additional Losses}
In addition to the losses suggested in the main paper, we incorporates several losses to better optimize the PIDNet as described below. For $\mathcal{L}_\mathrm{color}$ and $\mathcal{L}_\mathrm{ds}$, they were part of the baseline and showed a performance decrease upon removal. 
\paragraph{Edge-preserving loss.}
Motivated by \cite{godard2017unsupervised}, we used the gradient-based edge-preserving loss, to enforce the input and the novel view patches to preserve geometric properties. Using a weight term, $\omega(x)$, which already has been discussed in the paper, our edge-preserving loss on the predicted albedo can be formulated as:
\begin{equation} 
\begin{split}
     \mathcal{L}_{\mathrm{edge}} =\sum_{x' \in \mathcal{P}'} \omega(x)
     \|\partial(\hat{a}(x) - \hat{a}(x'))\|^2 ,
\end{split}
\label{eq:edge}
\end{equation}
where $\partial$ denotes the partial derivatives of the vertical and the horizontal directions, and $\mathcal{P'}$ denotes all the pixels in the target image. 

\paragraph{Chromaticity consistency loss.}
Similar to \cite{ye2022intrinsicnerf}, we adopted the chromaticity consistency loss to enforce the consistency between the input and novel view patches. It is formulated as:
\begin{equation}
  \mathcal{L}_{\mathrm{chrom}} = \sum_{x' \in \mathcal{P'}} \|\hat{ch}(x) - \hat{ch}(x')\|^2_2,
  \label{eq:ac}
\end{equation}
where $\hat{ch}(x)$ and $\hat{ch}(x')$ indicate the extracted albedo at $x$ and $x'$ from the novel view and the input view, respectively.
$\mathcal{P}$ denotes all the pixels in the novel view.

\paragraph{Intrinsic smoothness loss.}
Similar to the previous work~\cite{li2018learning} which uses various smoothness terms to give constraints to the network, we gave smoothness constraints to our patch-wise intrinsic decomposition network~(PIDNet). Our patch-wise intrinsic smoothness loss is formulated as:
\begin{equation}
\begin{split}
    \mathcal{L}_{\mathrm{pid}} = \sum_{x' \in \mathcal{P}'} \sum_{y \in \mathcal{N}(x')} \|\hat{a}(y)\!-\hat{a}(x')\|^2 ,
\end{split}
\label{eq:pid}
\end{equation}
where $y$ is one of the 4-neighbor adjacent pixels $\mathcal{N}(x')$ for $x'$, and $\mathcal{P}'$ denotes all the pixels of the target image.

\paragraph{Total loss functions. }
For each $x$ and $x'$ in the batch-wisely sampled input and the novel view patch, the total loss of our proposed framework is
\begin{equation}
\begin{split}
    \mathcal{L}_{\mathrm{total}}  =  & \lambda_{\mathrm{c}}\mathcal{L}_{\mathrm{color}} + \lambda_{\mathrm{a}}\mathcal{L}_{\mathrm{ac}} + \lambda_{\mathrm{dc}}\mathcal{L}_{\mathrm{dc}} + \lambda_{\mathrm{ds}}\mathcal{L}_{\mathrm{ds}} \\
    &+ \lambda_{\mathrm{e}}\mathcal{L}_{\mathrm{edge}} + \lambda_{\mathrm{pid}}\mathcal{L}_{\mathrm{pid}} + \lambda_{\mathrm{chrom}}\mathcal{L}_{\mathrm{chrom}}, 
\end{split}
\label{eq:total}
\end{equation}
where $\lambda_{\mathrm{c}}, \lambda_{\mathrm{a}}, \lambda_{\mathrm{dc}}, \lambda_{\mathrm{ds}}, \lambda_{\mathrm{edge}}, \lambda_{\mathrm{pid}}, \lambda_{\mathrm{chrom}}$ are weights parameters for each loss, respectively. In our experiment, losses are weighted as: $\lambda_{\mathrm{c}} = 1.0$, $\lambda_{\mathrm{a}} = 1.0$, $\lambda_{\mathrm{dc}} = 1.0$, $\lambda_{\mathrm{ds}} = 0.1$, $\lambda_{\mathrm{edge}} = 0.1$, $\lambda_{\mathrm{pid}} = 1.0$,  $\lambda_{\mathrm{chrom}} = 0.01$.

\begin{figure*}[hbt]
    \centering
    \begin{subfigure}[b]{1.0\textwidth}
    \centering
        \includegraphics[width=.15\linewidth]{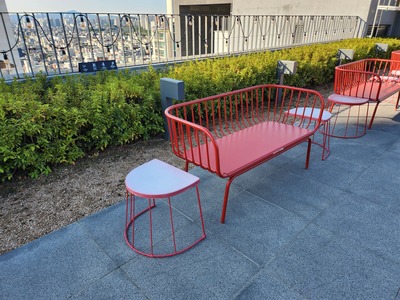}
        \includegraphics[width=.15\linewidth]{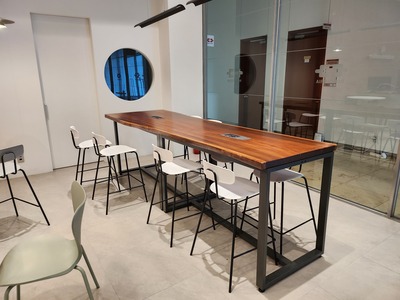}
        \includegraphics[width=.15\linewidth]{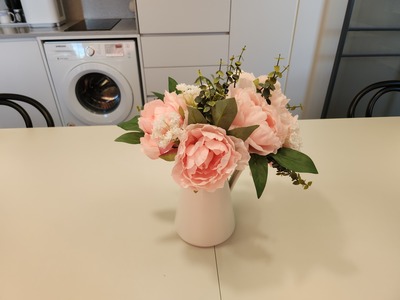}
        \includegraphics[width=.15\linewidth]{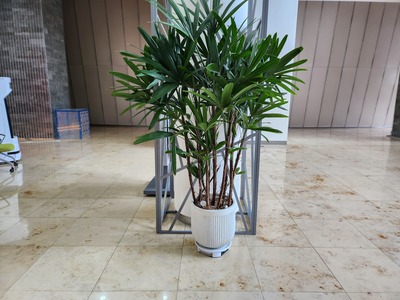}
        \includegraphics[width=.15\linewidth]{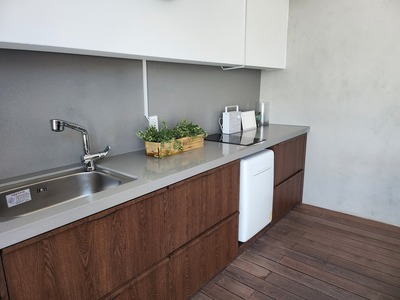}
        \vspace{-3pt}
     \end{subfigure}
     \begin{tabular}{>{\centering\arraybackslash}p{0.15\textwidth}>{\centering\arraybackslash}p{0.12\textwidth}>{\centering\arraybackslash}p{0.15\textwidth}>{\centering\arraybackslash}p{0.12\textwidth}>{\centering\arraybackslash}p{0.15\textwidth}}
    \scriptsize Bench  & \scriptsize Cafe & \scriptsize Flower  & \scriptsize Houseplant & \scriptsize Kitchen  \\
    \end{tabular}
    \begin{subfigure}[b]{1.0\textwidth}
    \centering
       \includegraphics[width=.15\linewidth]{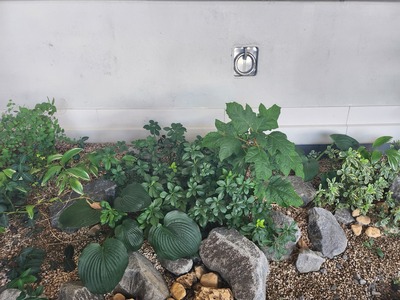}
       \includegraphics[width=.15\linewidth]{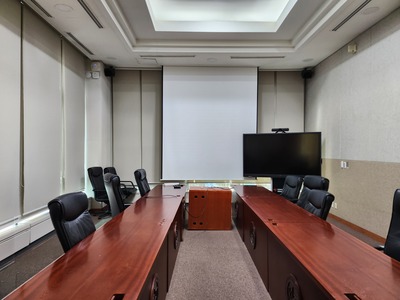}
       \includegraphics[width=.15\linewidth]{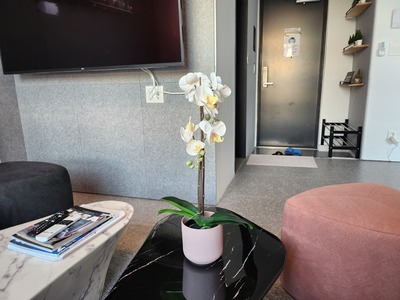}
       \includegraphics[width=.15\linewidth]{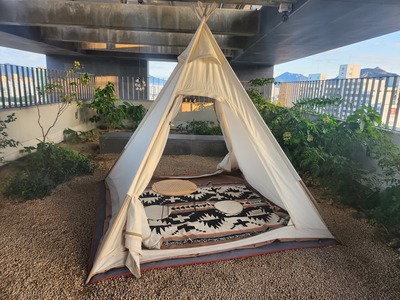}
       \includegraphics[width=.15\linewidth]{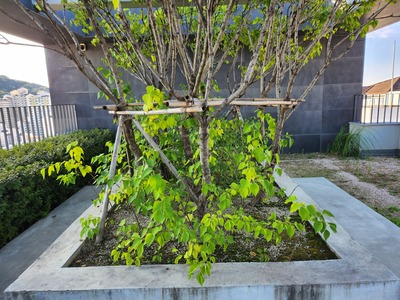}
        \vspace{-3pt}
    \end{subfigure}
    \begin{tabular}{>{\centering\arraybackslash}p{0.15\textwidth}>{\centering\arraybackslash}p{0.12\textwidth}>{\centering\arraybackslash}p{0.15\textwidth}>{\centering\arraybackslash}p{0.12\textwidth}>{\centering\arraybackslash}p{0.15\textwidth}}
    \scriptsize Leaves  & \scriptsize Room & \scriptsize Table & \scriptsize Tent & \scriptsize Tree \\
    \end{tabular}
    \caption{
        \textbf{NeRF Extreme dataset.} This dataset is newly built to provide multi-view images under varying illumination, which can be used to train and evaluate the robust NeRF. Exemplified images are selected from their test sets with mild-lighting conditions.}
    \label{fig:scenelist} 
\end{figure*}

\begin{figure}[hbt]
    \centering
        \includegraphics[width=\linewidth]{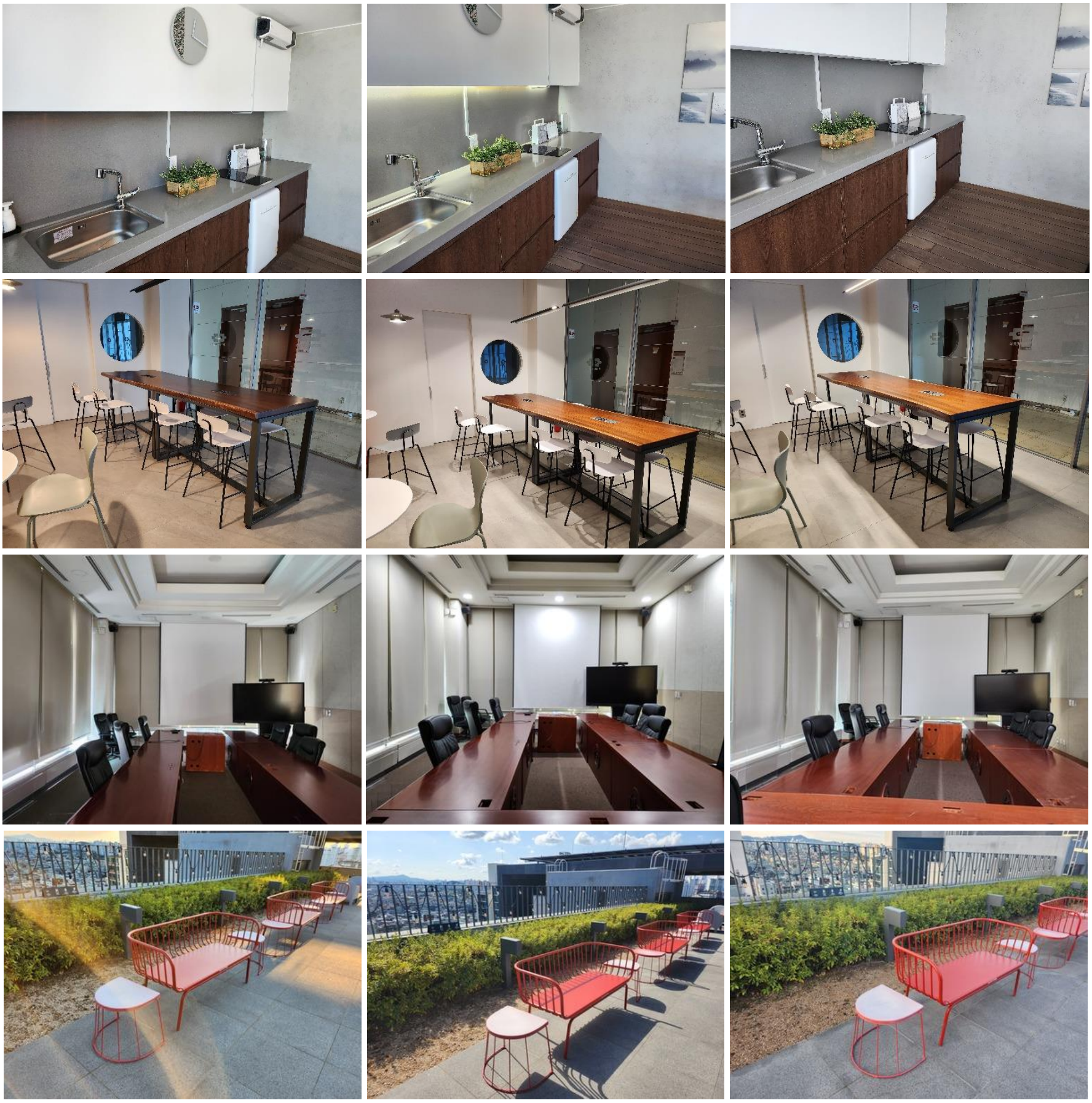}
    \caption{
    \textbf{Examples of inputs sampled from NeRF Extreme.}
   Sampled scenes with varying illumination from the `Kitchen', `Cafe', `Room', and `Bench', respectively. 
    }
    \label{fig:inputs_extreme} 
\end{figure}

\subsection{Baseline Implementations}
\paragraph{NeRF under varying illumination.}
Since the official implementation of NeRF-W~\cite{nerfinthewild} is not available, we used a well-known implementation of the paper~\cite{queianchen_nerf} for the comparison. In the case of Ha-NeRF~\cite{HaNeRF}, the official codes were used for the synthesis and hallucination. All data were trained using the provided hyperparameters with a batch size of $1024$.

\paragraph{Few-shot NeRF.}
For both RegNeRF~\cite{niemeyer2022regnerf} and FreeNeRF~\cite{yang2023freenerf}, the official JAX~\cite{jax2018github} implementations of the papers was used for comparison. The configurations for the LLFF~\cite{mildenhall2019local} were used for the training, for their same data characteristics. 

\paragraph{Few-shot NeRF under varying illumination.}
In the case of NeROIC, the official implementation was used for comparison. Note that the original code involves noisy camera pose correction during optimization. For a comparison with aligned test poses, we set the training option `optimize$\_$camera' to be false. Furthermore, all the front-facing scenes were treated as images paired with masks that hold true for every pixel.

\section{More Details on Datasets}  
\subsection{NeRF Extreme}
Our dataset consists of 10 scenes captured both indoors and outdoors. Fig.~\ref{fig:scenelist} shows the representative images for each scene of our dataset.
All scenes contain 40 images each, with a resolution of 3,000 x 4,000. Among these, 30 images are split for the training set, while the remaining 10 are allocated for the test set. The training images encompass at least three distinct lighting conditions, effectively simulating a highly practical environment characterized by varying illumination inputs. Exemplified images in our dataset can be found at the end of the material.
Notably, the `Bench,' `Tent,' and `Tree' scenes constitute the challenging subset. These scenes feature difficult aspects such as hard shadows, unbounded depths (sky), and repeated textures like leaves, which can pose challenges for NeRF-based methods. 
The synthesized results on the `Cafe', `Kitchen', and `Flower' scenes are reported in the paper with $3$ input view settings. Other synthesized results can be found in the next section. 

\begin{figure}
  \centering
  \includegraphics[width=\linewidth]{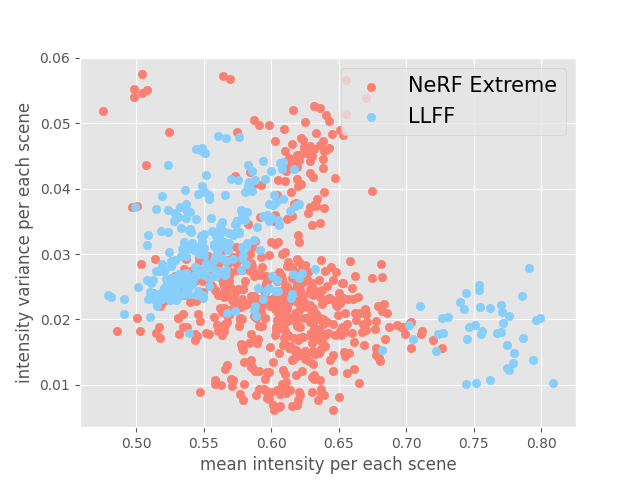}
   \caption{\textbf{Intensity distribution of our NeRF Extreme and LLFF~\cite{mildenhall2019local}.} Our dataset shows a larger variance in per-image lighting intensity distribution than LLFF. Lighting intensities are obtained from the shading images extracted by \cite{das2022pie}.}
   \label{fig:graph}
\end{figure}

\paragraph{Lighting variations.}
In order to verify the illumination diversity of our dataset, we extract per-image intensity distribution from shading images, which are obtained by the state-of-the-art intrinsic decomposition framework~\cite{das2022pie}. A shading image is suitable for evaluating the illumination diversity of a dataset since it provides environment-dependent information about the image. Fig.~\ref{fig:graph} shows an intensity distribution of ours and existing LLFF~\cite{mildenhall2019local}, which has similar dataset characteristics. Each dot in the distribution indicates per-image intensity characteristics. Our dataset shows more scattered distribution compared to LLFF, indicating a larger illumination diversity.

\begin{figure}[t]
    \centering
        \includegraphics[width=\linewidth]{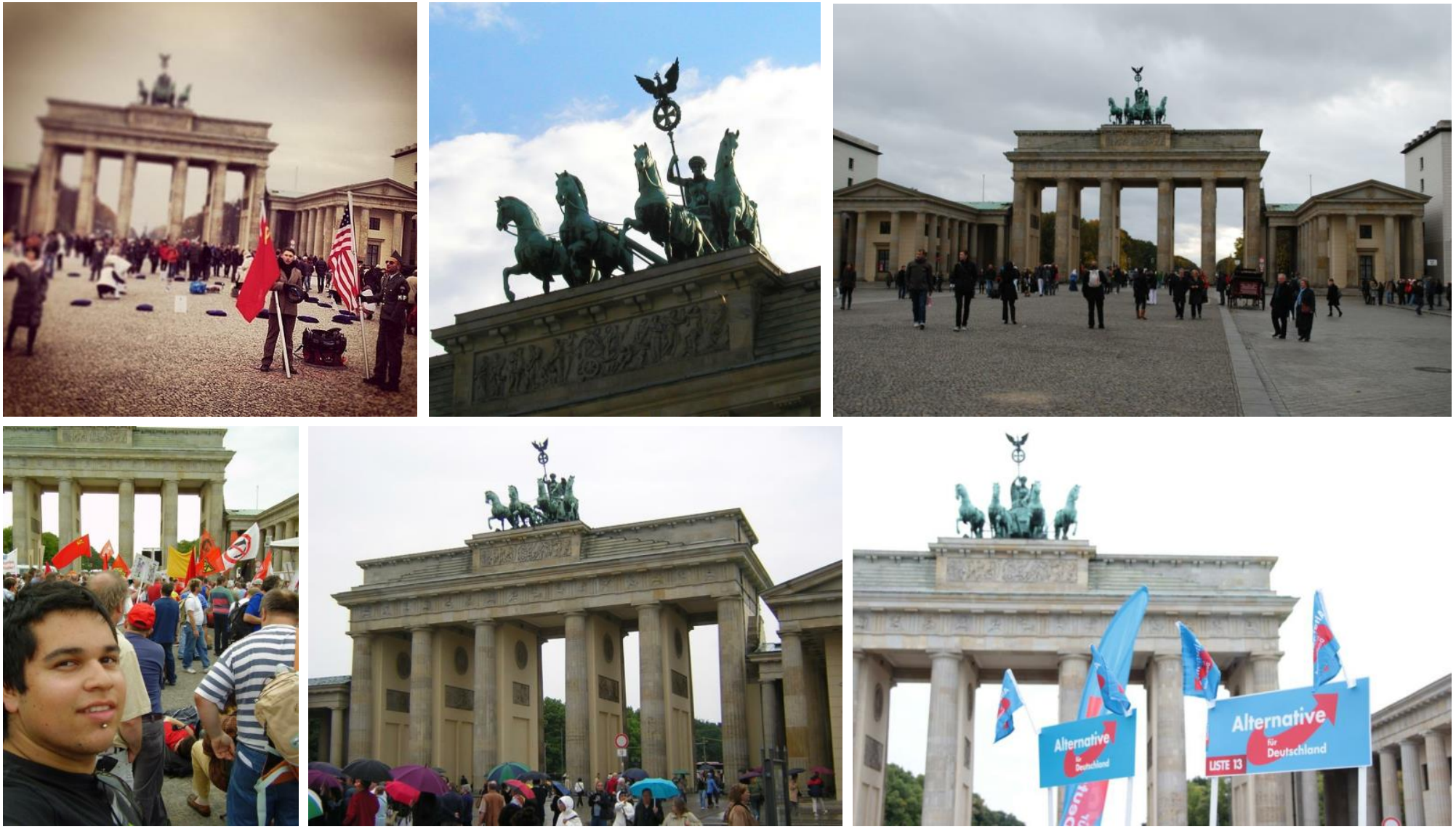}
    \caption{
    \textbf{Examples of images randomly sampled from Phototourism~\cite{snavely2006photo}.}
   The sampled scenes exhibit a high degree of randomness in terms of camera pose, field-of-view, occlusions, and depth bounds.
    }
    \label{fig:tourism_random} 
\end{figure}

\subsection{Phototourism $\mathbf{F}^3$}
For the evaluation of few-shot view synthesis under varying illumination, about $12$ to $15$ images were selected to construct a training set of Phototourism $\mathbf{F}^3$, as illustrated in Fig.~\ref{fig:inputs_tourism}. Note that selected image ids are based on the description of the previous work~\cite{HaNeRF}, instead of the original dataset~\cite{snavely2006photo}. The problem of misalignment in image ids is introduced in the official implementation of Ha-NeRF. 

\paragraph{The rationale behind building a frontal-facing subset.}
In this paragraph, we demonstrate the need for constructing a frontal-facing subset of Phototourism, for the evaluation of few-shot NeRF under varying illumination. Fig.~\ref{fig:tourism_random} shows examples of images randomly sampled from the original dataset. It can be easily found that the exampled photos have a high degree of randomness in various aspects, taken by random cameras with varying camera intrinsics. In specific cases other than a substantial number of input images are provided, a few-shot view synthesis using the dataset is impossible to handle. To mitigate this problem, we constructed a subset of Phototourism, as Phototourism $\mathbf{F}^3$~(frontal-facing few-shot), which has frontal-facing scenes with at least similar field-of-views and depth bounds.

\section{Additional Experimental Results}
\subsection{Additional Ablation Studies}
\paragraph{Ablation on learned knowledge priors.}
Some of the previous works have attempted to utilize prior knowledge for few-shot view synthesis. 
To enrich the evaluation of the proposed method, we compare our method with PixelNeRF~\cite{yu2021pixelnerf} and MVSNeRF~\cite{chen2021mvsnerf}, which utilize CNN feature extractor and 3D cost volume, respectively. Tab.~\ref{tab:prior} presents the performance comparison using 3 input views of the 'Cafe' scene in the NeRF Extreme benchmark. The difference in SSIM and LPIPS metrics indicates that PixelNeRF and MVSNeRF yield cluttered view synthesis results when presented with inputs featuring varying illumination. 
In the third row, we present results by replacing our proposed albedo consistency with perceptual loss, which might offer positive effects in managing varying illumination inputs. However, perceptual loss prioritizes visual feasibility over handling illumination, leading to a degradation in performance.

\begin{table} [t]
    \centering
    \resizebox{0.8\linewidth}{!}
    {\begin{tabular}{l|ccc}
\toprule
  & SSIM $\uparrow$ & LPIPS $\downarrow$ & Abs Rel $\downarrow$ \\ 
  \midrule
PixelNeRF~[CVPR'21] \ & 0.30 & 0.71 & 0.96 \\
MVSNeRF~[ICCV'21] \ & 0.42 & 0.46 & \bf{0.50} \\
\midrule
Ours~(albedo $\rightarrow$ perc.) \ & 0.42 & 0.41 & 0.91 \\
\textbf{Ours} \ & \textbf{0.48} & \bf{0.38} & \textbf{0.51} \\
\bottomrule
\end{tabular}
}
    \vspace{-2mm}
    \caption{   
    \textbf{Additional ablation study on the learned priors.}}
    \label{tab:prior}
\end{table}

\paragraph{Ablation on patch size.}
Tab.~\ref{tab:patch} shows additional ablation on the patch size of PIDNet with the `Brandenburg Gate' scene, given 3 input views in the Phototourism $\mathbf{F}^3$. Incorporating large patch-wise projections can lead to poor geometry, as it may include pixels with different depths within the same patch. This can result in inaccurate reconstructions, making it essential to strike a balance in determining an appropriate patch size to avoid the inclusion of unrelated surfaces during cross-view regularization.
Conversely, incorporating a small patch size may hinder PIDNet's performance, lacking enough context. As a result, $S_{\mathrm{patch}} = 32 \times 32$ yields optimal results.

\begin{table} [hbt!]
    \centering
    \resizebox{0.8\linewidth}{!}
    {\begin{tabular}{l|ccc}
\toprule
  & SSIM $\uparrow$ & LPIPS $\downarrow$ & Abs Rel $\downarrow$   \\ \midrule
$S_{\mathrm{patch}} = 64 \times 64 $ \ & 0.45 & 0.49 & 0.87\\
$S_{\mathrm{patch}} = 32 \times 32 $ \ & \textbf{0.56} & \textbf{ 0.36} & \textbf{0.78} \\
$S_{\mathrm{patch}} = 16 \times 16 $ \ & 0.48 & 0.46 & 0.79 \\ 
$S_{\mathrm{patch}} = 8 \times 8 $ \ & 0.50 & 0.44 & 0.78 \\ 
\bottomrule
\end{tabular}
}
    \vspace{-2mm}
    \caption{
    \textbf{Ablation study on patch size.} Incorporating unsuitable patch sizes leads to performance degradation.}
    \label{tab:patch}
\end{table}

\subsection{Video Synthesis Results}
We also provide video result comparisons with the baselines, as supplementary material. All the synthesized videos were rendered by the model trained with three input views. As emphasized in the paper, our ExtremeNeRF shows superior performance in recovering geometry details compared to the other methods, given only sparse inputs taken under unconstrained illumination. Some transient components with unreliable depth - a person working in front of the `Brandenburg Gate', for example, and illumination changed within frames, can be deleted if the proper techniques are utilized. However, properly handling transient components and relighting the scenes given only a sparse number of inputs is still a challenging problem.

\subsection{Novel View Synthesis on Object-centric Dataset}
In this subsection, we provide additional experimental results on two types of object-centric datasets. 

\paragraph{NeROIC.}
We compared our model with NeROIC~\cite{kuang2022neroic} on their `TV' dataset, following the evaluation strategy provided by the paper. Note that datasets provided by NeRD~\cite{boss2021nerd} and NeROIC, including TV datasets, are object-centric scenes that provide 360 scenes of the given object. We randomly selected $10$ images from the training set to formulate the situation of few-shot view synthesis under varying illumination. 
Since our projective-transformation-based consistency regularization can be better applied to situations where input scenes depict the same points as many as they can, these 360 object-centric scenes are not favorable inputs for us, especially in a few-shot setting.
Moreover, the dataset is provided with foreground masks and white backgrounds, which lack the global contexts necessary for successful intrinsic decomposition. 

\begin{figure}[t]
    \centering
     \begin{tabular}{>{\centering\arraybackslash}p{.26\linewidth}>{\centering\arraybackslash}p{.30\linewidth}>{\centering\arraybackslash}p{.30\linewidth}}
     \scriptsize NeROIC-Full & \scriptsize NeROIC-Geom. & \scriptsize \textbf{ExtremeNeRF (Ours)}\\
      \end{tabular}
         \centering
        \includegraphics[width=\linewidth]{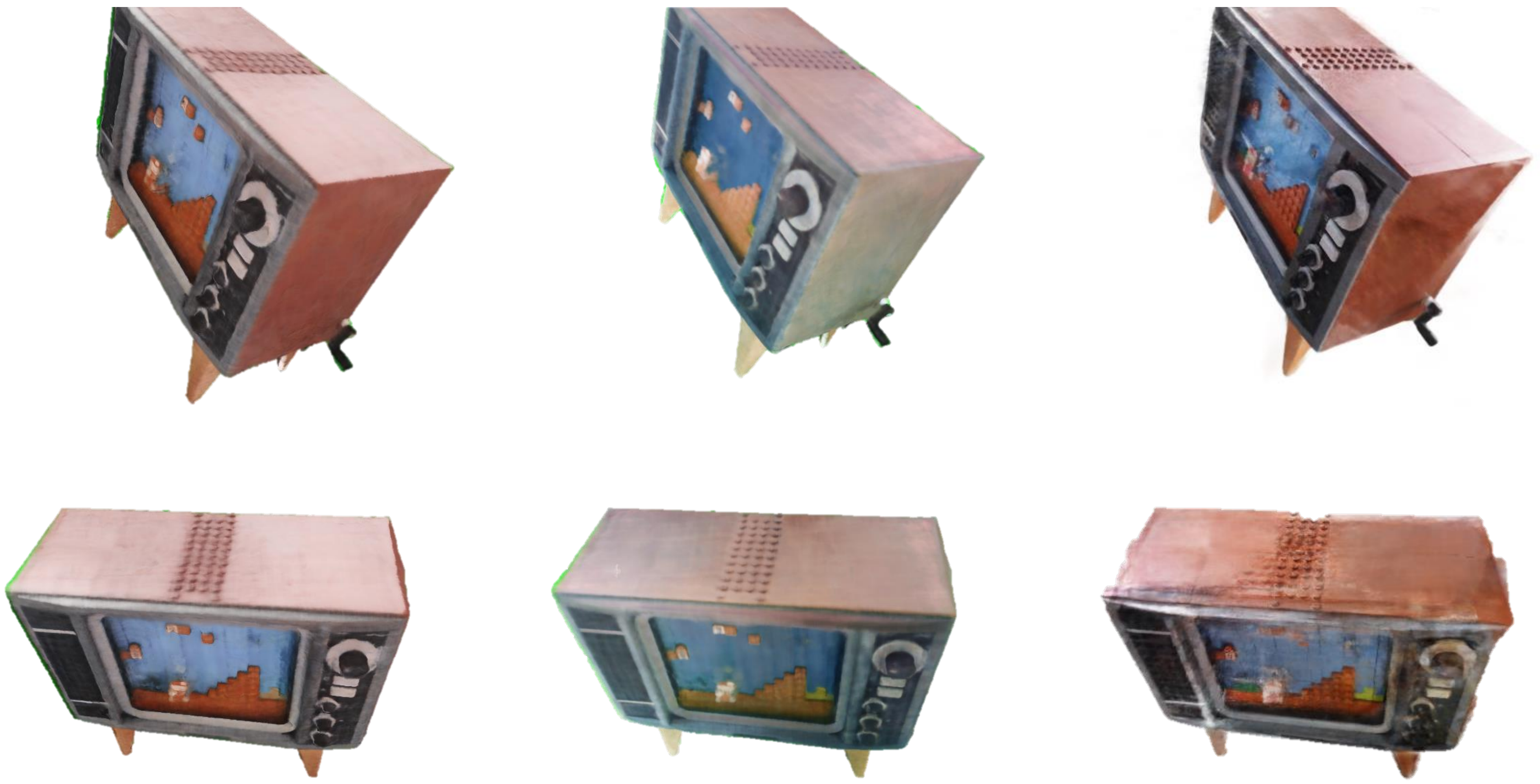}
    \caption{
        \textbf{Few-shot view synthesis results on the `TV' scene of NeROIC Dataset~\cite{kuang2022neroic}.}
    In unfavorable environments where global contexts are lacking and pixel-wise correspondences are weakened, our ExtremeNeRF shows hindered yet reasonable synthesized results.}
    \label{fig:tv}
\end{figure}

Table~\ref{tab:tv} and Fig.~\ref{fig:tv} present both qualitative and quantitative results on the `TV' dataset. Our proposed method faces challenges due to the lack of global contexts and weak pixel-to-pixel correspondences; however, it still manages to produce credible results. The comparison of LPIPS scores in Table~\ref{tab:tv} highlights that our proposed method achieves reasonable outcomes even with challenging, object-centric inputs that lack backgrounds.

\begin{table} [hbt]
    \centering
    \resizebox{0.8\linewidth}{!}
    {\begin{tabular}{l|ccc}
\toprule
  & SSIM $\uparrow$ & LPIPS $\downarrow$ & Abs Rel $\downarrow$   \\ \midrule
NeROIC-Geom. \ & \textbf{0.66} & \textbf{0.40} & N/A\\
NeROIC-Full. \ & 0.64 & 0.48 & N/A\\
\textbf{ExtremeNeRF~(Ours)} \ & 0.58 & \textbf{0.40} & N/A\\
\bottomrule
\end{tabular}
}
    \caption{
    \textbf{Quantitative comparison on the  `TV' scene of NeROIC Dataset~\cite{kuang2022neroic}.}
   Our proposed method shows a compatible LPIPS score compared to the baseline, despite the unfavorable inputs.}
    \label{tab:tv} 
\end{table}

\paragraph{Light-varying DTU.}
DTU~\cite{DTU} consists of images taken under structured cameras and light sources. There exist seven number of lighting variations per a scene. Previous works which deal with view synthesis under consistent illumination~\cite{mildenhall2020nerf, dietnerf, yu2021pixelnerf, srf, barron2021mip, niemeyer2022regnerf, CLIPnerf, infonerf, chen2021mvsnerf, xu2022sinnerf} have used the dataset with fixed mild lighting condition. 
Note that DTU is not a dataset with 360 scenes as NeROIC. DTU consists of frontal-facing scenes with objects located at the center of the context-discarded backgrounds. 

In this paper, we randomly chose lighting conditions for each scene, for the experiments on few-shot view synthesis under varying illumination. Following the evaluation protocol used by the previous works~\cite{yu2021pixelnerf, niemeyer2022regnerf}, we used scan IDs (8, 21, 30, 31, 34, 38, 40, 41, 45, 55, 63, 82, 103, 110, 114) as the dataset, while using image IDs (25, 22, 28) as inputs. In the cases of the lighting condition, (4, 1, 5) were used for (25, 22, 28) images, respectively. All images were used in a resolution of $300 \times 400$, following the evaluation protocols of the previous work~\cite{niemeyer2022regnerf}. 

\begin{figure*}
    \centering
    \begin{tabular}{>{\centering\arraybackslash}p{0.32\textwidth}>{\centering\arraybackslash}p{0.28\textwidth}>{\centering\arraybackslash}p{0.32\textwidth}}
      \scriptsize \textbf{NeRF under varying-illumination} & \scriptsize \textbf{Few-shot NeRF} & \scriptsize \textbf{Few-shot NeRF under varying-illumination}\\
    \end{tabular}
    \begin{tabular}{>{\centering\arraybackslash}p{0.13\textwidth}>{\centering\arraybackslash}p{0.15\textwidth}>{\centering\arraybackslash}p{0.15\textwidth}>{\centering\arraybackslash}p{0.14\textwidth}>
    {\centering\arraybackslash}p{0.15\textwidth}>
    {\centering\arraybackslash}p{0.13\textwidth}}
    \scriptsize NeRF-W~(CVPR'21) & \scriptsize Ha-NeRF~(CVPR'22) & \scriptsize RegNeRF~(CVPR'22) & \scriptsize FreeNeRF~(CVPR'23)  & 
    \scriptsize NeROIC-Geom.~(SIG'22) &
    \scriptsize  \textbf{ExtremeNeRF~(Ours)}\\
    \end{tabular}
     \begin{subfigure}[b]{1.0\textwidth} 
             \centering
        \includegraphics[width=\linewidth]{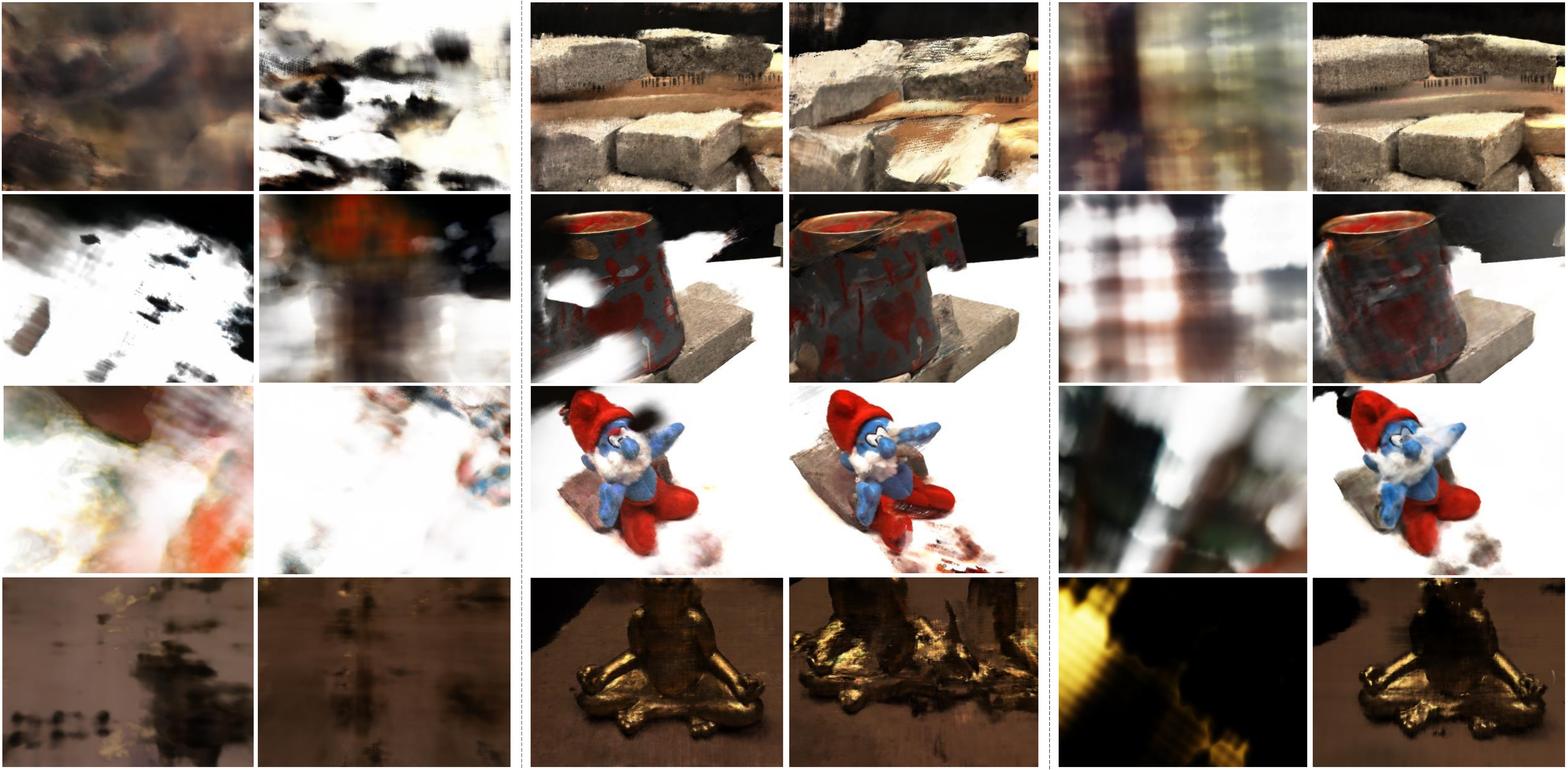}
     \end{subfigure}

    \caption{
    \textbf{Additional qualitative comparison on DTU~\cite{DTU}.}
    A synthesized novel view of the `Scan38', `Scan41', `Scan82', and `Scan110' scene~(from top to bottom), generated by the baselines and our proposed method.
    }
    \label{fig:dtu} 
\end{figure*}

Fig.~\ref{fig:dtu} and Tab.~\ref{tab:dtu} present both qualitative and quantitative comparisons of our method with baselines on the dataset. The baselines, except for few-shot NeRF, exhibit substantial distortions in their performance. RegNeRF~\cite{niemeyer2022regnerf} and FreeNeRF~\cite{yang2023freenerf} display reasonable synthesis performance; however, the geometry details suffer due to the illumination variations in the inputs. In most cases, our proposed method either outperforms or ranks second-best in performance. However, the rate of performance improvement has been reduced compared to experiments on other datasets that feature frontal-facing scenes with robust global contexts. This reduction in improvement points to the significance of global contexts within a scene for successful intrinsic decomposition. All metrics were calculated without masks.

\subsection{Additional Results on Phototourism}
\begin{figure*}
    \centering
     \begin{tabular}{>{\centering\arraybackslash}p{0.32\textwidth}>{\centering\arraybackslash}p{0.28\textwidth}>{\centering\arraybackslash}p{0.32\textwidth}}
      \scriptsize \textbf{NeRF under varying-illumination} & \scriptsize \textbf{Few-shot NeRF} & \scriptsize \textbf{Few-shot NeRF under varying-illumination}\\
    \end{tabular}
    \begin{tabular}{>{\centering\arraybackslash}p{0.13\textwidth}>{\centering\arraybackslash}p{0.15\textwidth}>{\centering\arraybackslash}p{0.15\textwidth}>{\centering\arraybackslash}p{0.14\textwidth}>
    {\centering\arraybackslash}p{0.15\textwidth}>
    {\centering\arraybackslash}p{0.13\textwidth}}
    \scriptsize NeRF-W~(CVPR'21) & \scriptsize Ha-NeRF~(CVPR'22) & \scriptsize RegNeRF~(CVPR'22) & \scriptsize FreeNeRF~(CVPR'23)  & 
    \scriptsize NeROIC-Geom.~(SIG'22) &
    \scriptsize  \textbf{ExtremeNeRF~(Ours)}\\
    \end{tabular}
     \begin{subfigure}[b]{1.0\textwidth} 
         \centering
        \includegraphics[width=\linewidth]{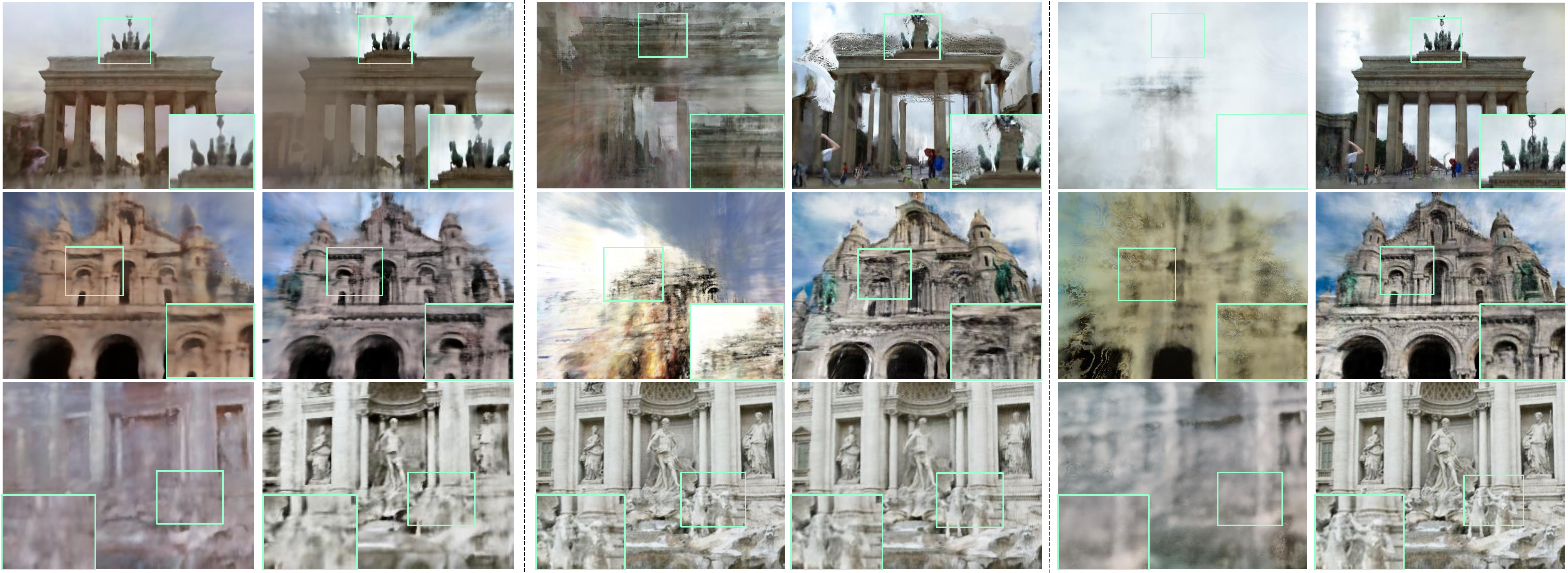}
     \end{subfigure}
    \caption{
    \textbf{Additional qualitative comparison on Phototourism $\mathbf{F}^3$ benchmark.}
    }
    \label{fig:supp_tourism} 
\end{figure*}

In Fig.~\ref{fig:supp_tourism}, we provide additional experimental results on Phototourism $\mathbf{F}^3$ in a $3$ view setting. In most cases, our proposed method shows superior performance in synthesizing novel views given sparse inputs taken under varying illumination. Especially, in the `Brandenburg Gate' scene~(Top), our proposed method shows plausible geometry details while other baselines suffer from distorted geometries~\cite{niemeyer2022regnerf, yang2023freenerf}, or floating artifacts~\cite{nerfinthewild, HaNeRF, kuang2022neroic}. In the `Trevi Fountain' scene, RegNeRF seems to have a compatible performance compared to us. However, Tab.~3 in the paper, proves that RegNeRF is hindered by the varying illumination input in other test poses, showing a $0.14$ lower SSIM score compared to ours.

\subsection{Additional Results on NeRF Extreme}
\begin{figure*}
    \centering
    \begin{tabular}{>{\centering\arraybackslash}p{0.32\textwidth}>{\centering\arraybackslash}p{0.28\textwidth}>{\centering\arraybackslash}p{0.32\textwidth}}
      \scriptsize \textbf{NeRF under varying-illumination} & \scriptsize \textbf{Few-shot NeRF} & \scriptsize \textbf{Few-shot NeRF under varying-illumination}\\
    \end{tabular}
    \begin{tabular}{>{\centering\arraybackslash}p{0.13\textwidth}>{\centering\arraybackslash}p{0.15\textwidth}>{\centering\arraybackslash}p{0.15\textwidth}>{\centering\arraybackslash}p{0.14\textwidth}>
    {\centering\arraybackslash}p{0.15\textwidth}>
    {\centering\arraybackslash}p{0.13\textwidth}}
    \scriptsize NeRF-W~(CVPR'21) & \scriptsize Ha-NeRF~(CVPR'22) & \scriptsize RegNeRF~(CVPR'22) & \scriptsize FreeNeRF~(CVPR'23)  & 
    \scriptsize NeROIC-Geom.~(SIG'22) &
    \scriptsize  \textbf{ExtremeNeRF~(Ours)}\\
    \end{tabular}
     \begin{subfigure}[b]{1.0\textwidth} 
             \centering
        \includegraphics[width=\linewidth]{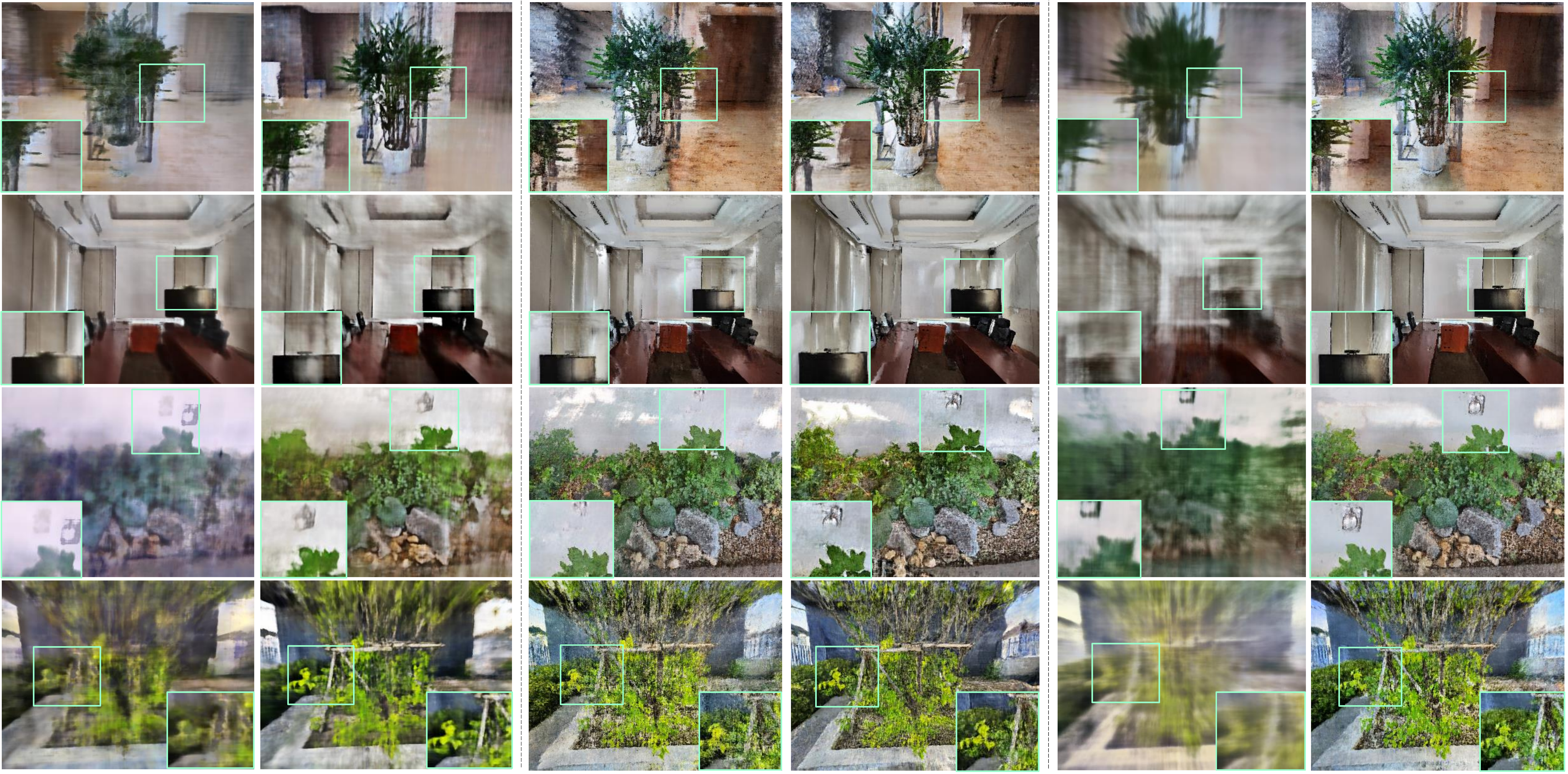}
     \end{subfigure}

    \caption{
    \textbf{Additional qualitative comparison on NeRF Extreme benchmark.}
    A synthesized novel view of the `Houseplant', `Room', `Leaves', and `Tree' scene~(from top to bottom), generated by the baselines and our proposed method.
    }
    \label{fig:supp_extreme} 
\end{figure*}

Fig.~\ref{fig:supp_extreme} and Tab.~\ref{tab:supp_extreme} demonstrate additional qualitative and quantitative results on NeRF Extreme benchmark. In most cases, our proposed method shows the best performance in synthesizing novel views with fine geometry details. The light-green boxes emphasize the reasonable geometry details of our results, while other results from the baselines are hindered by the challenging inputs. However, when it comes to the 'Tree' scene, which is part of the challenging subset of the dataset, all the methods, including ours, yield unsatisfactory results. The presence of unbounded depth (sky) in the scene adversely affects performance. The presence of unreasonable synthesis results and low metric scores prove the fact that few-shot NeRF under varying illumination is an exceedingly demanding task that still requires significant progress and advancements.

\section{NeRF Extreme Benchmark Examples}
\begin{figure*}
    \centering
    \begin{subfigure}[b]{1.0\textwidth}
         \centering
        \includegraphics[width=.18\linewidth]{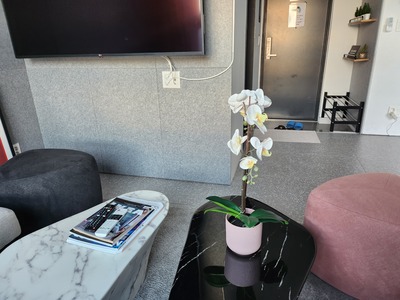}
        \includegraphics[width=.18\linewidth]{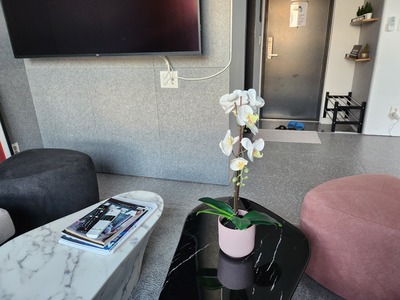}
        \includegraphics[width=.18\linewidth]{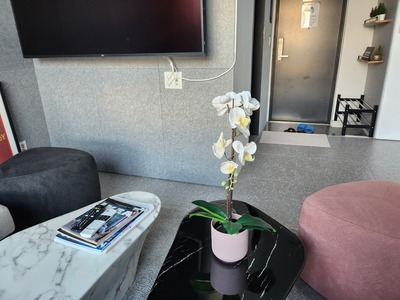}
        \includegraphics[width=.18\linewidth]{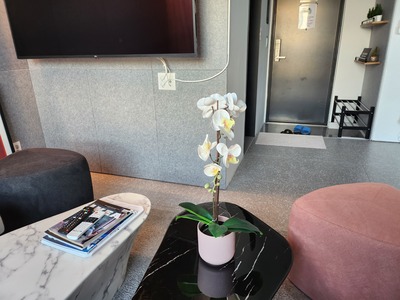}
        \includegraphics[width=.18\linewidth]{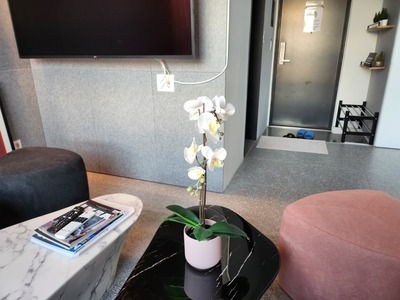}
        \vspace{.125cm}
    \end{subfigure}
    \begin{subfigure}[b]{1.0\textwidth}
    \centering
        \includegraphics[width=.18\linewidth]{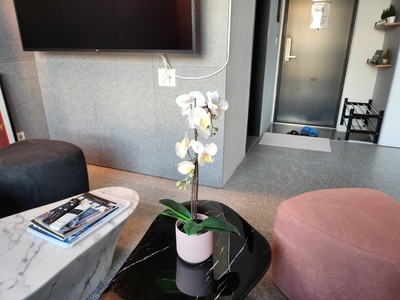}
        \includegraphics[width=.18\linewidth]{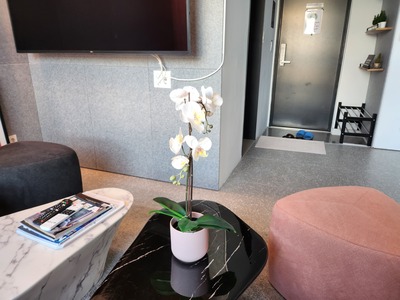}
        \includegraphics[width=.18\linewidth]{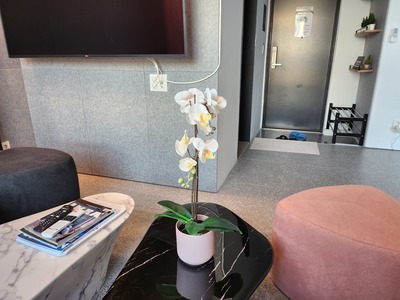}
        \includegraphics[width=.18\linewidth]{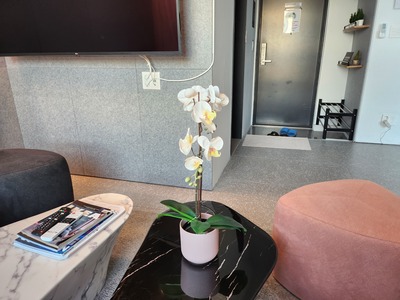}
        \includegraphics[width=.18\linewidth]{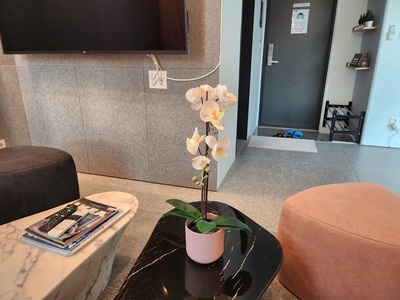}
        \vspace{.125cm}
    \end{subfigure}
    \begin{subfigure}[b]{1.0\textwidth}
    \centering
        \includegraphics[width=.18\linewidth]{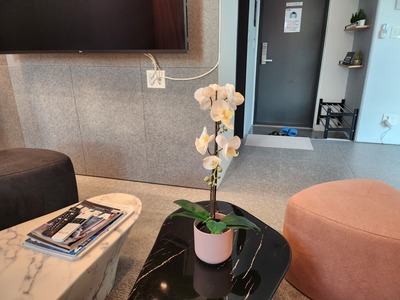}
        \includegraphics[width=.18\linewidth]{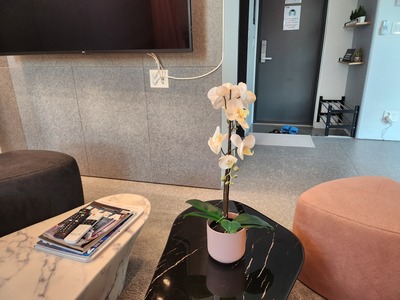}
        \includegraphics[width=.18\linewidth]{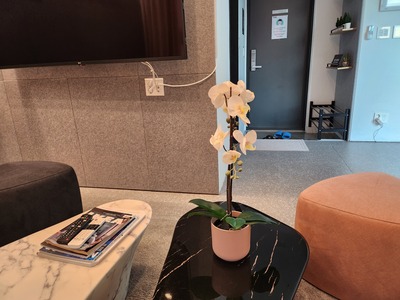}
        \includegraphics[width=.18\linewidth]{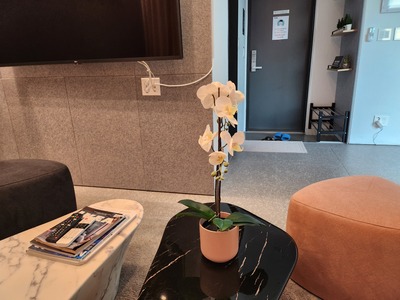}
        \includegraphics[width=.18\linewidth]{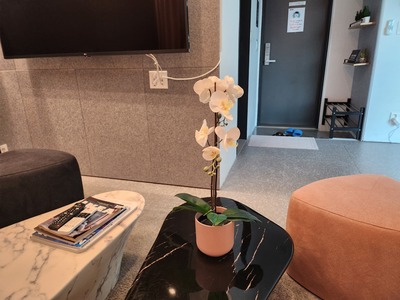}
        \vspace{.125cm}
    \end{subfigure}
    \begin{subfigure}[b]{1.0\textwidth}
    \centering
        \includegraphics[width=.18\linewidth]{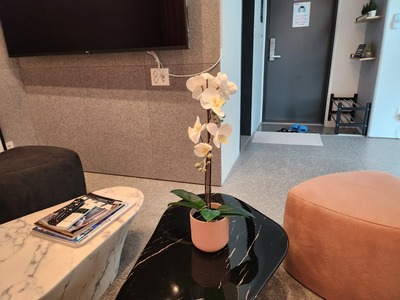}
        \includegraphics[width=.18\linewidth]{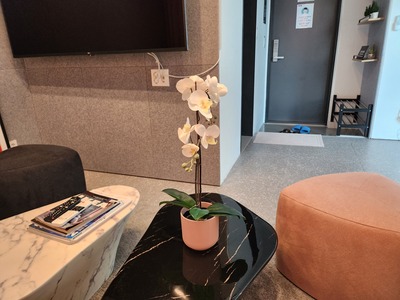}
        \includegraphics[width=.18\linewidth]{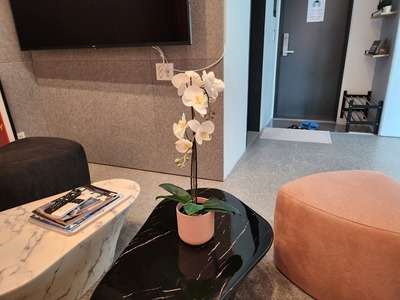}
        \includegraphics[width=.18\linewidth]{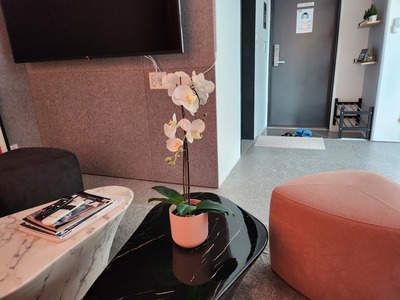}
        \includegraphics[width=.18\linewidth]{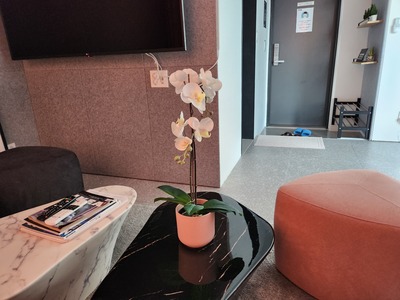}
        \vspace{.125cm}
    \end{subfigure}
    \begin{subfigure}[b]{1.0\textwidth}
    \centering
        \includegraphics[width=.18\linewidth]{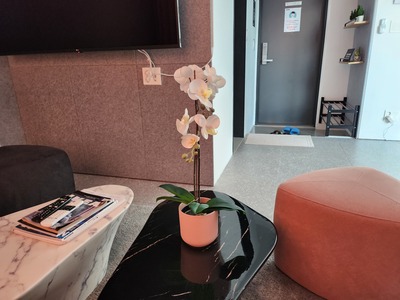}
        \includegraphics[width=.18\linewidth]{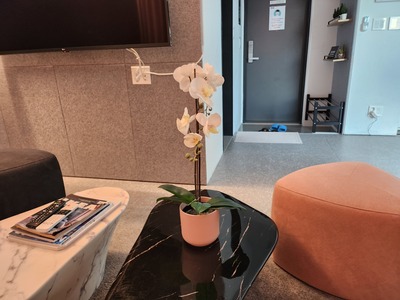}
        \includegraphics[width=.18\linewidth]{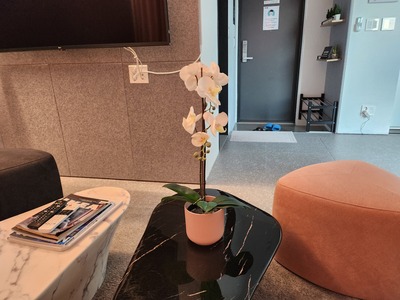}
        \includegraphics[width=.18\linewidth]{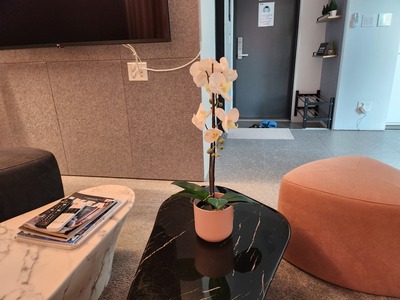}
        \includegraphics[width=.18\linewidth]{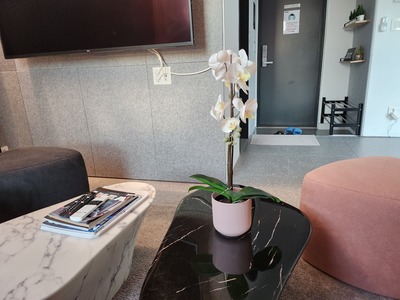}
        \vspace{.125cm}
    \end{subfigure}
    \begin{subfigure}[b]{1.0\textwidth}
    \centering
        \includegraphics[width=.18\linewidth]{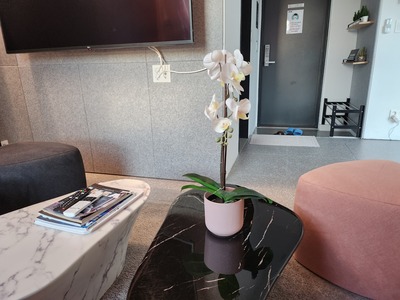}
        \includegraphics[width=.18\linewidth]{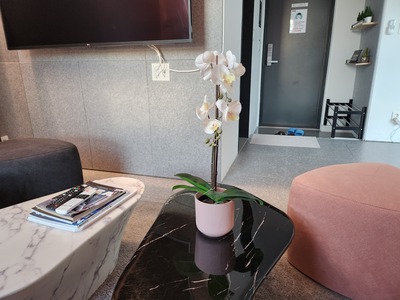}
        \includegraphics[width=.18\linewidth]{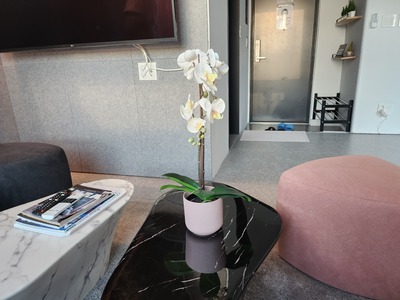}
        \includegraphics[width=.18\linewidth]{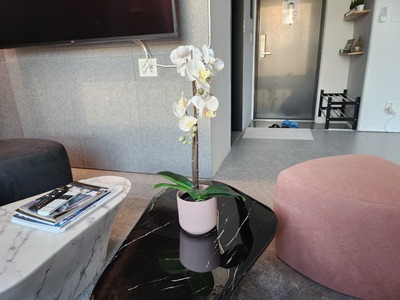}
        \includegraphics[width=.18\linewidth]{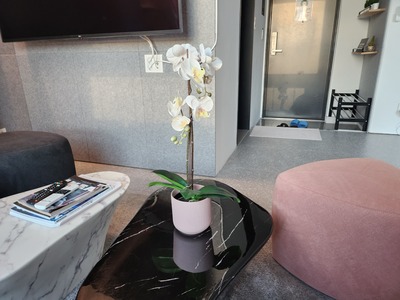}
        \vspace{.125cm}
    \end{subfigure}
    \begin{subfigure}[b]{1.0\textwidth}
    \centering
        \includegraphics[width=.18\linewidth]{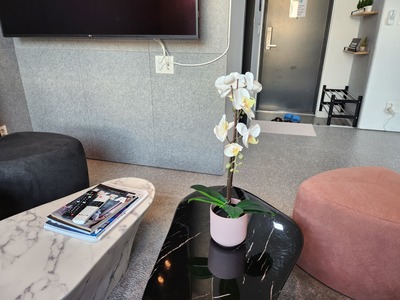}
        \includegraphics[width=.18\linewidth]{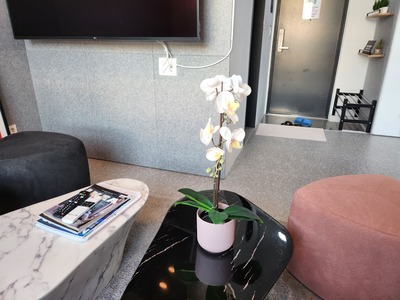}
        \includegraphics[width=.18\linewidth]{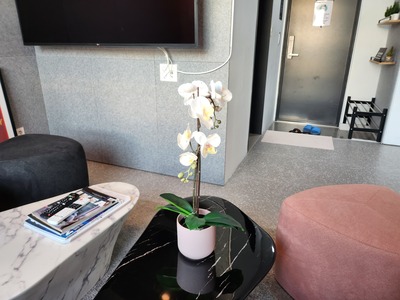}
        \includegraphics[width=.18\linewidth]{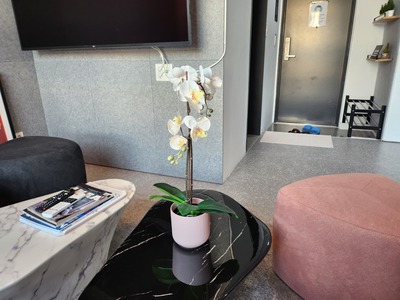}
        \includegraphics[width=.18\linewidth]{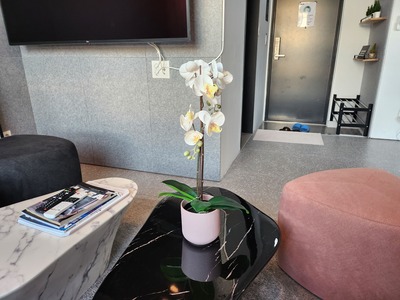}
        \vspace{.125cm}
    \end{subfigure}
    \begin{subfigure}[b]{1.0\textwidth}
    \centering
        \includegraphics[width=.18\linewidth]{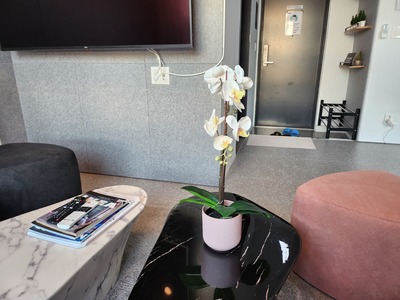}
        \includegraphics[width=.18\linewidth]{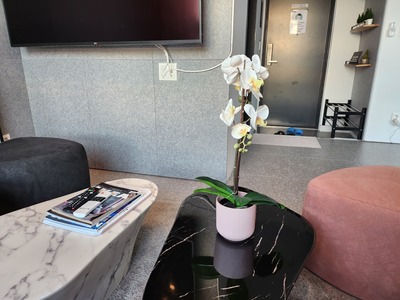}
        \includegraphics[width=.18\linewidth]{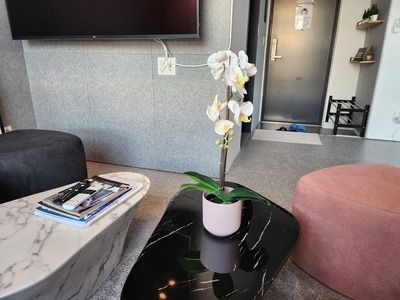}
        \includegraphics[width=.18\linewidth]{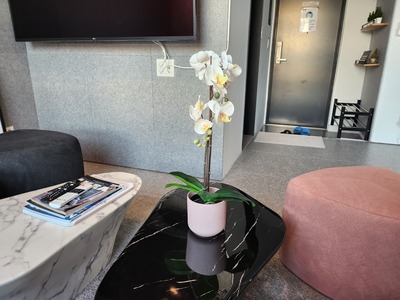}
        \includegraphics[width=.18\linewidth]{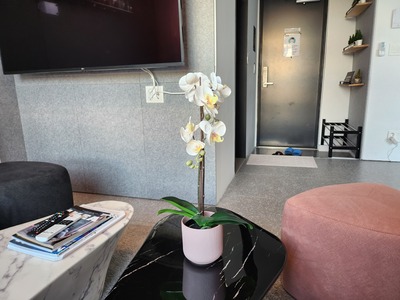}
    \end{subfigure}
    \caption{
        \textbf{Illumination variation samples of the \textit{table} scene in NeRF Extreme. }}
    \label{fig:tableall} 
\end{figure*}

\begin{figure*}
    \centering
    \begin{subfigure}[b]{1.0\textwidth}
         \centering
        \includegraphics[width=.18\linewidth]{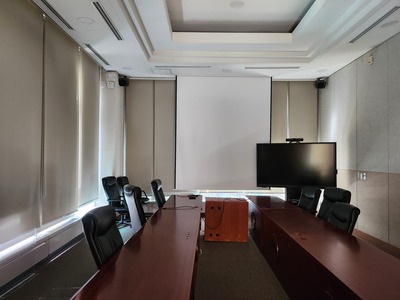}
        \includegraphics[width=.18\linewidth]{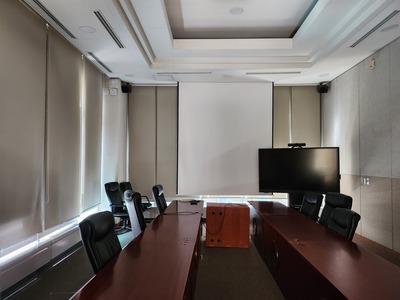}
        \includegraphics[width=.18\linewidth]{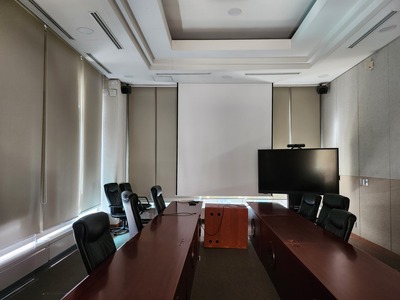}
        \includegraphics[width=.18\linewidth]{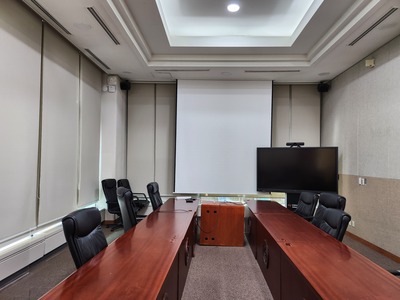}
        \includegraphics[width=.18\linewidth]{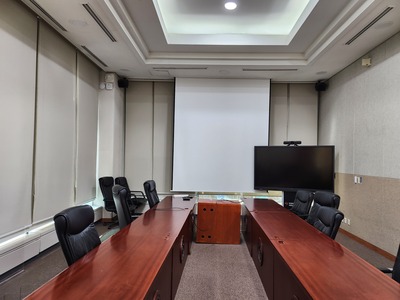}
        \vspace{.125cm}
    \end{subfigure}
    \begin{subfigure}[b]{1.0\textwidth}
    \centering
        \includegraphics[width=.18\linewidth]{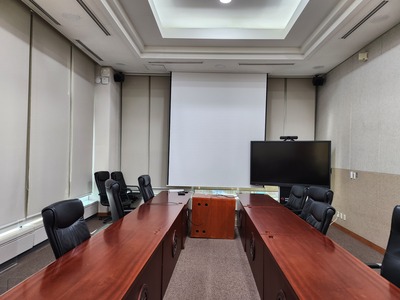}
        \includegraphics[width=.18\linewidth]{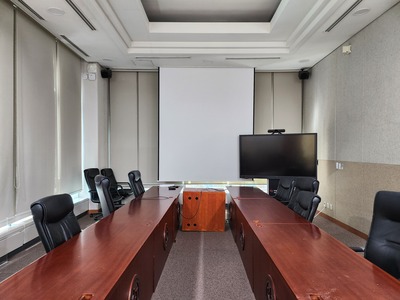}
        \includegraphics[width=.18\linewidth]{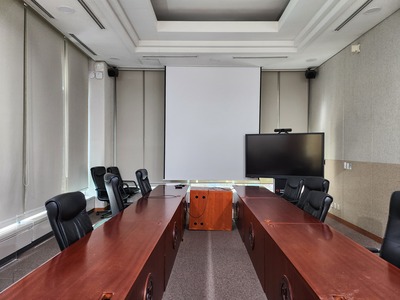}
        \includegraphics[width=.18\linewidth]{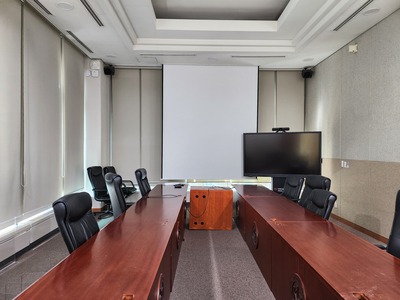}
        \includegraphics[width=.18\linewidth]{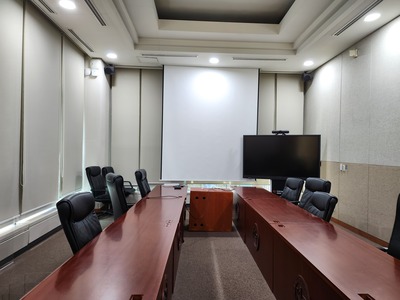}
        \vspace{.125cm}
    \end{subfigure}
    \begin{subfigure}[b]{1.0\textwidth}
    \centering
        \includegraphics[width=.18\linewidth]{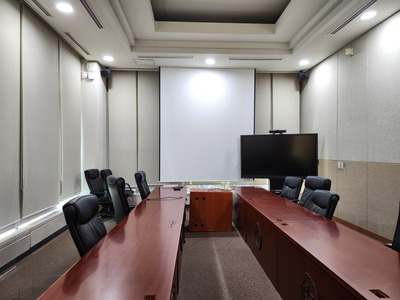}
        \includegraphics[width=.18\linewidth]{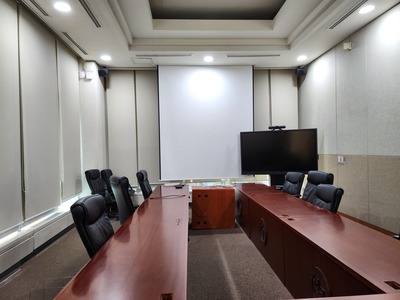}
        \includegraphics[width=.18\linewidth]{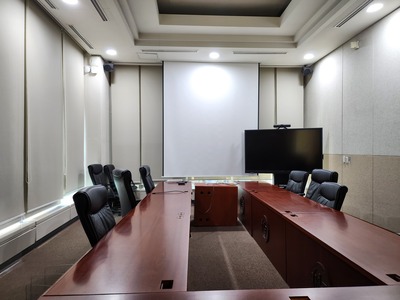}
        \includegraphics[width=.18\linewidth]{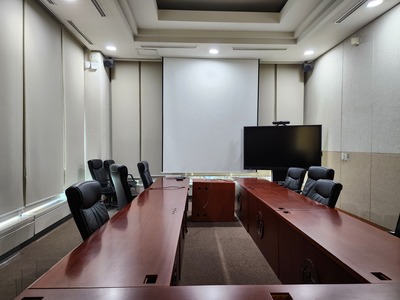}
        \includegraphics[width=.18\linewidth]{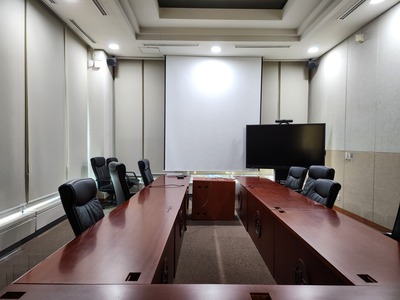}
        \vspace{.125cm}
    \end{subfigure}
    \begin{subfigure}[b]{1.0\textwidth}
    \centering
        \includegraphics[width=.18\linewidth]{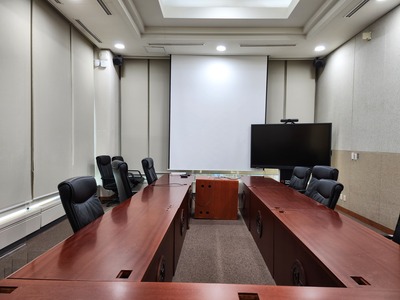}
        \includegraphics[width=.18\linewidth]{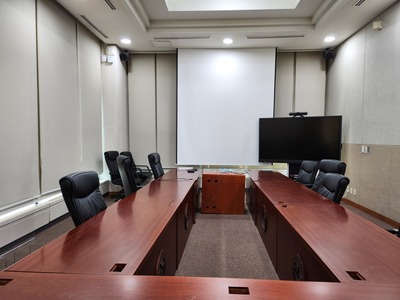}
        \includegraphics[width=.18\linewidth]{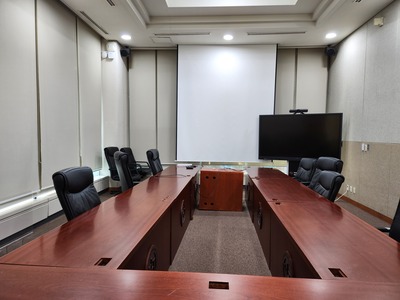}
        \includegraphics[width=.18\linewidth]{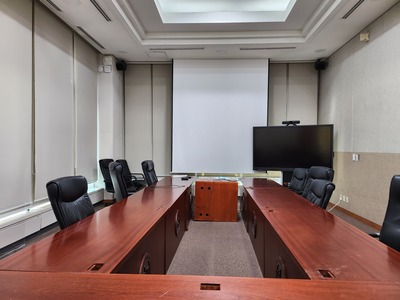}
        \includegraphics[width=.18\linewidth]{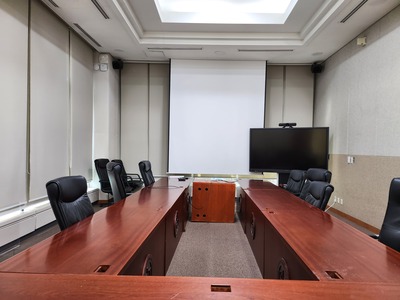}
        \vspace{.125cm}
    \end{subfigure}
    \begin{subfigure}[b]{1.0\textwidth}
    \centering
        \includegraphics[width=.18\linewidth]{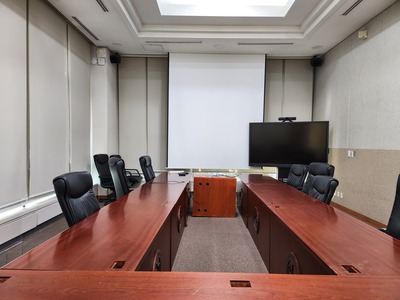}
        \includegraphics[width=.18\linewidth]{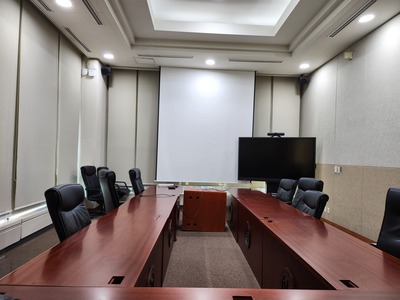}
        \includegraphics[width=.18\linewidth]{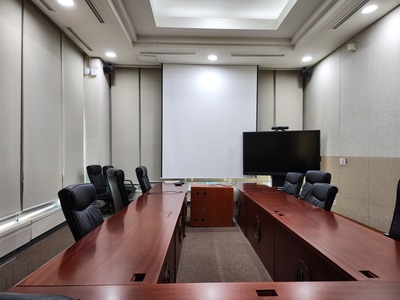}
        \includegraphics[width=.18\linewidth]{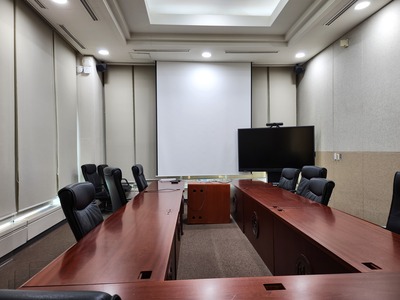}
        \includegraphics[width=.18\linewidth]{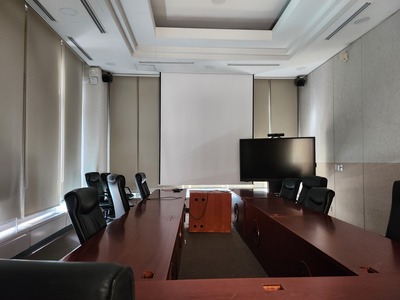}
        \vspace{.125cm}
    \end{subfigure}
    \begin{subfigure}[b]{1.0\textwidth}
    \centering
        \includegraphics[width=.18\linewidth]{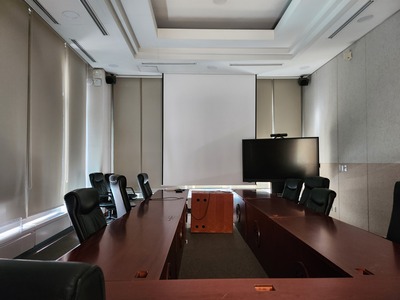}
        \includegraphics[width=.18\linewidth]{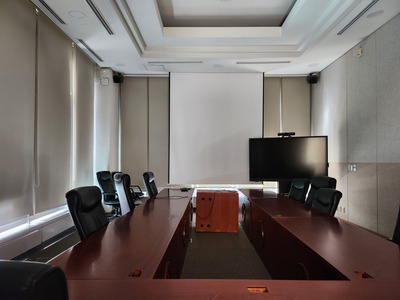}
        \includegraphics[width=.18\linewidth]{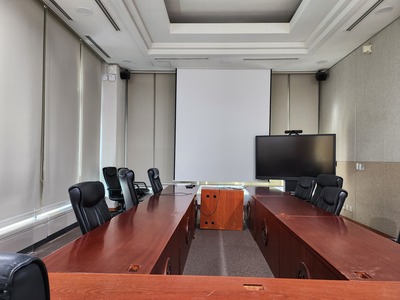}
        \includegraphics[width=.18\linewidth]{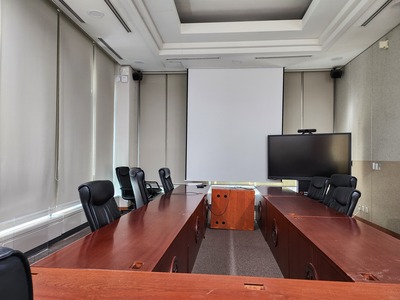}
        \includegraphics[width=.18\linewidth]{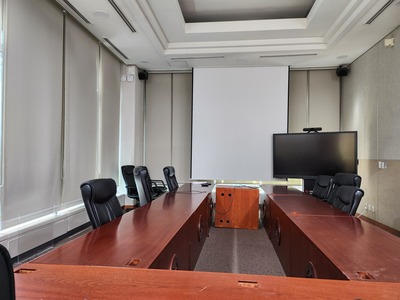}
        \vspace{.125cm}
    \end{subfigure}
    \begin{subfigure}[b]{1.0\textwidth}
    \centering
        \includegraphics[width=.18\linewidth]{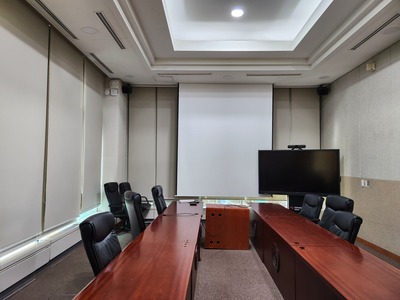}
        \includegraphics[width=.18\linewidth]{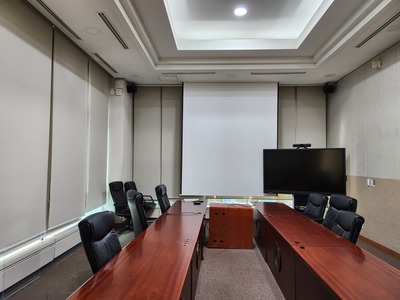}
        \includegraphics[width=.18\linewidth]{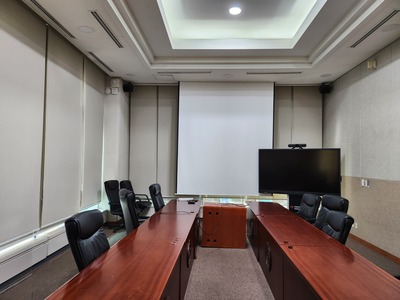}
        \includegraphics[width=.18\linewidth]{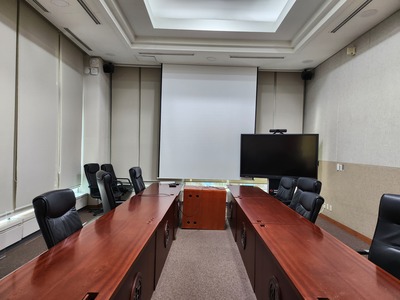}
        \includegraphics[width=.18\linewidth]{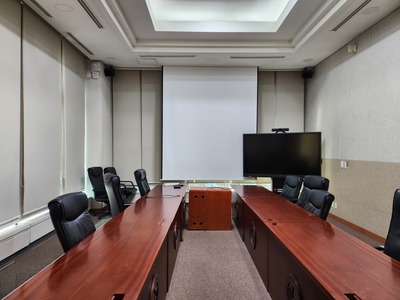}
        \vspace{.125cm}
    \end{subfigure}
    \begin{subfigure}[b]{1.0\textwidth}
    \centering
        \includegraphics[width=.18\linewidth]{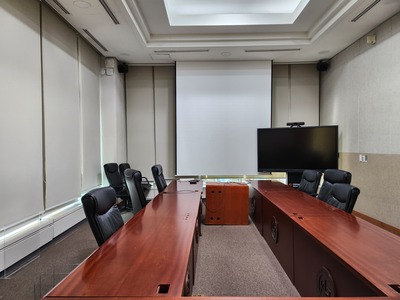}
        \includegraphics[width=.18\linewidth]{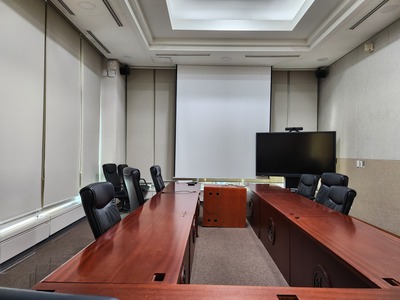}
        \includegraphics[width=.18\linewidth]{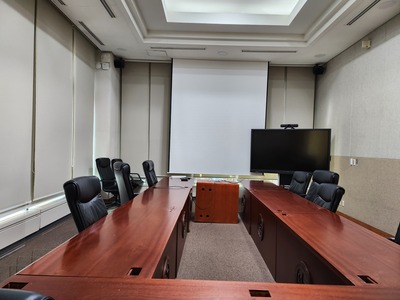}
        \includegraphics[width=.18\linewidth]{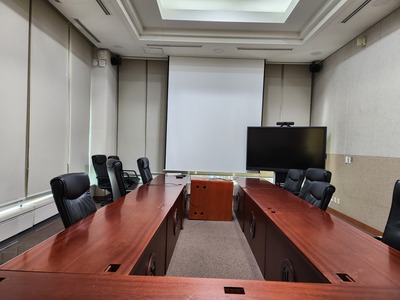}
        \includegraphics[width=.18\linewidth]{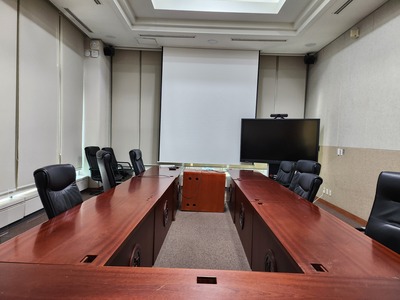}
    \end{subfigure}
    \caption{
        \textbf{Illumination variation samples of the \textit{room} scene in NeRF Extreme. }}
    \label{fig:roomall}
\end{figure*}

\begin{figure*}
    \centering
    \begin{subfigure}[b]{1.0\textwidth}
         \centering
        \includegraphics[width=.18\linewidth]{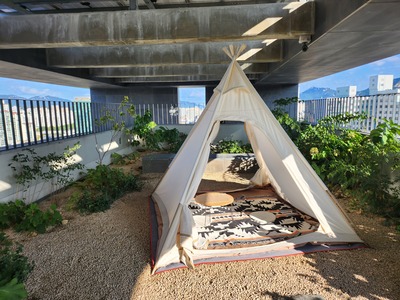}
        \includegraphics[width=.18\linewidth]{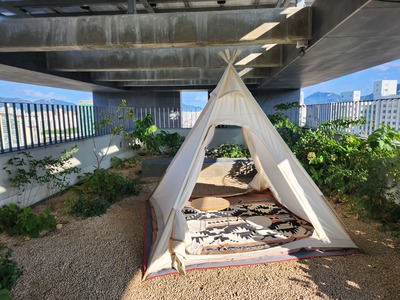}
        \includegraphics[width=.18\linewidth]{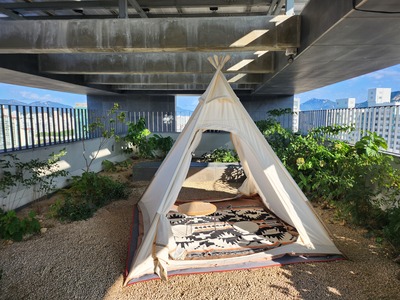}
        \includegraphics[width=.18\linewidth]{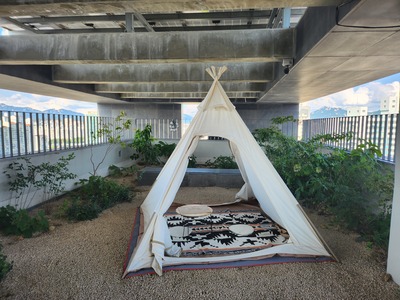}
        \includegraphics[width=.18\linewidth]{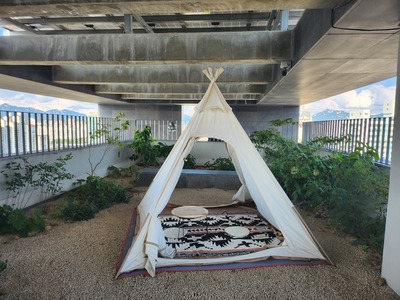}
        \vspace{.125cm}
    \end{subfigure}
    \begin{subfigure}[b]{1.0\textwidth}
    \centering
        \includegraphics[width=.18\linewidth]{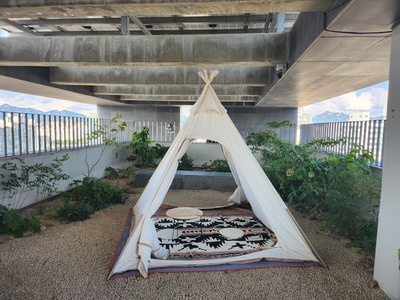}
        \includegraphics[width=.18\linewidth]{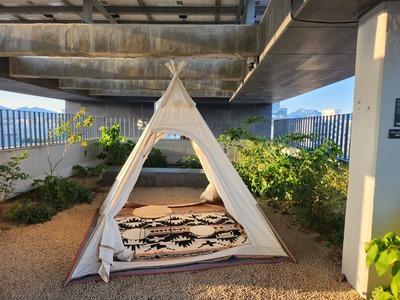}
        \includegraphics[width=.18\linewidth]{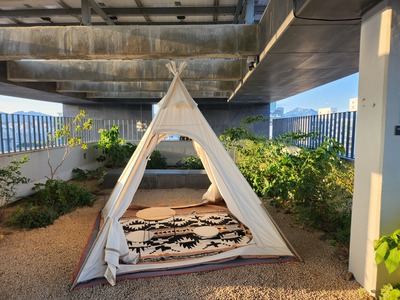}
        \includegraphics[width=.18\linewidth]{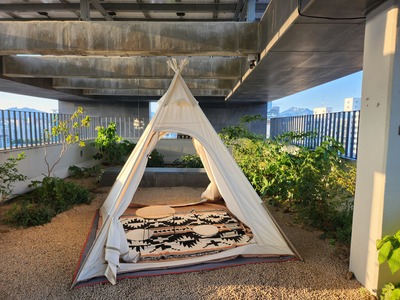}
        \includegraphics[width=.18\linewidth]{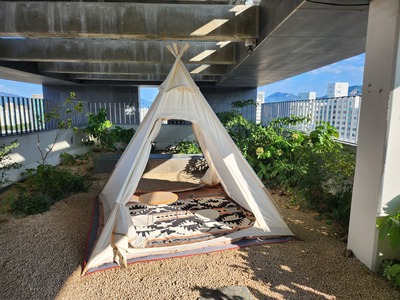}
        \vspace{.125cm}
    \end{subfigure}
    \begin{subfigure}[b]{1.0\textwidth}
    \centering
        \includegraphics[width=.18\linewidth]{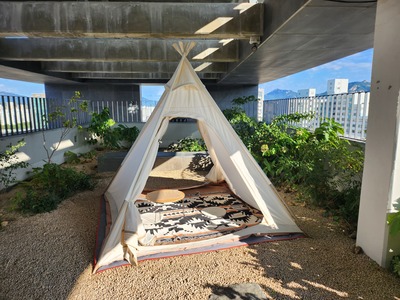}
        \includegraphics[width=.18\linewidth]{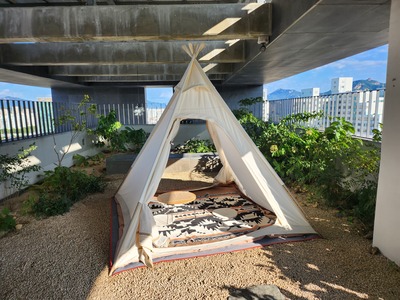}
        \includegraphics[width=.18\linewidth]{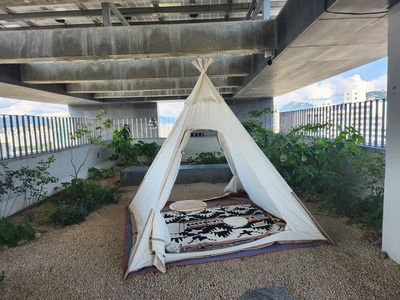}
        \includegraphics[width=.18\linewidth]{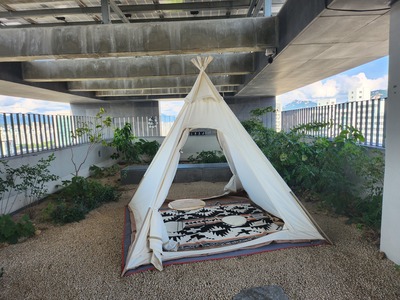}
        \includegraphics[width=.18\linewidth]{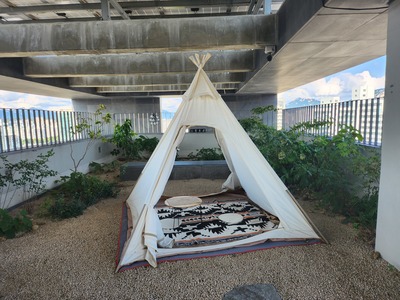}
        \vspace{.125cm}
    \end{subfigure}
    \begin{subfigure}[b]{1.0\textwidth}
    \centering
        \includegraphics[width=.18\linewidth]{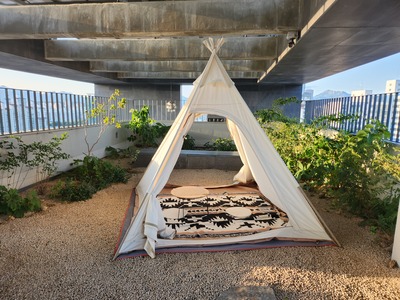}
        \includegraphics[width=.18\linewidth]{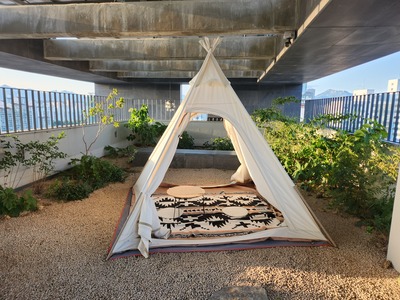}
        \includegraphics[width=.18\linewidth]{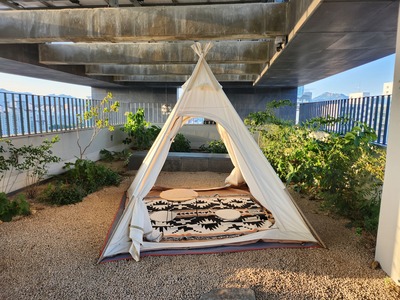}
        \includegraphics[width=.18\linewidth]{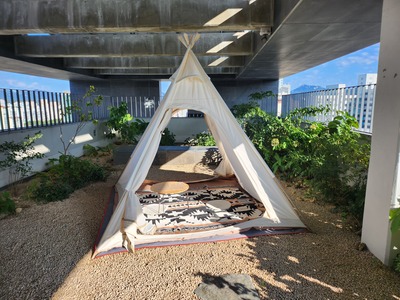}
        \includegraphics[width=.18\linewidth]{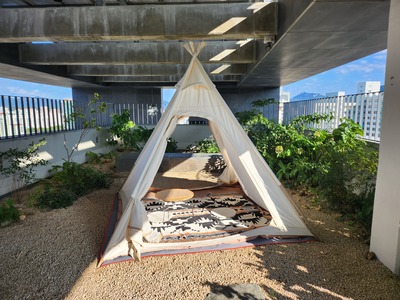}
        \vspace{.125cm}
    \end{subfigure}
    \begin{subfigure}[b]{1.0\textwidth}
    \centering
        \includegraphics[width=.18\linewidth]{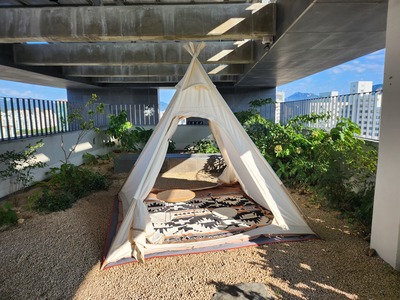}
        \includegraphics[width=.18\linewidth]{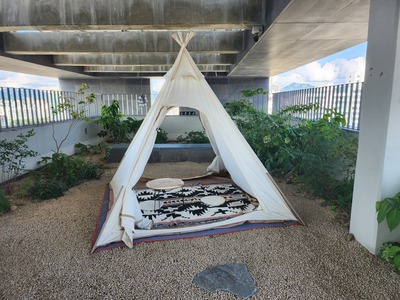}
        \includegraphics[width=.18\linewidth]{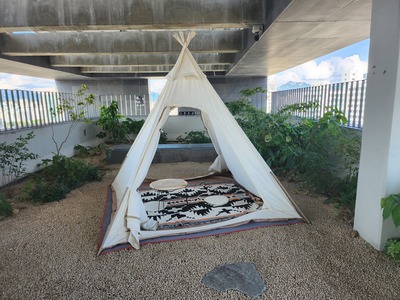}
        \includegraphics[width=.18\linewidth]{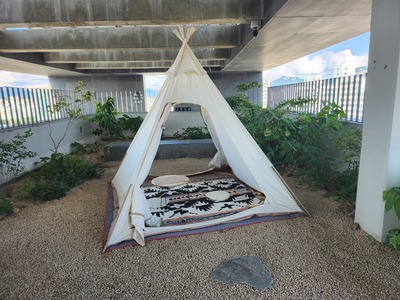}
        \includegraphics[width=.18\linewidth]{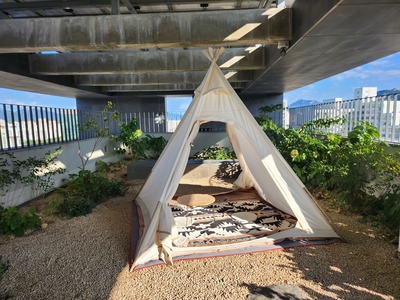}
        \vspace{.125cm}
    \end{subfigure}
    \begin{subfigure}[b]{1.0\textwidth}
    \centering
        \includegraphics[width=.18\linewidth]{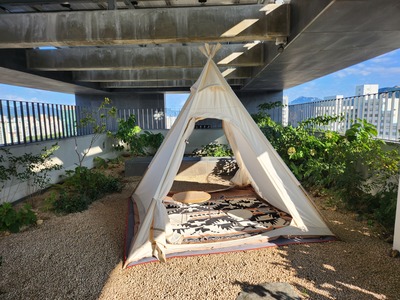}
        \includegraphics[width=.18\linewidth]{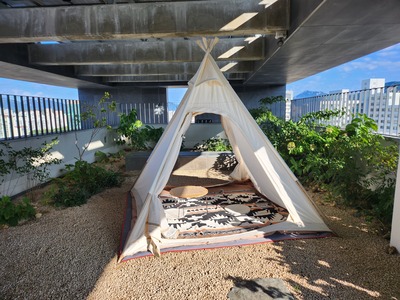}
        \includegraphics[width=.18\linewidth]{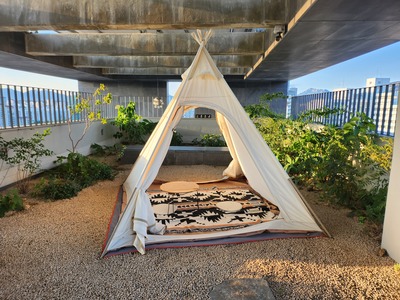}
        \includegraphics[width=.18\linewidth]{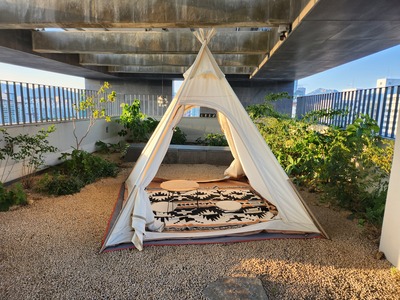}
        \includegraphics[width=.18\linewidth]{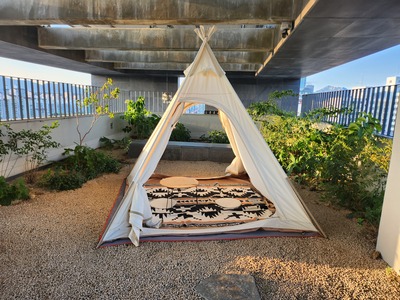}
        \vspace{.125cm}
    \end{subfigure}
    \begin{subfigure}[b]{1.0\textwidth}
    \centering
        \includegraphics[width=.18\linewidth]{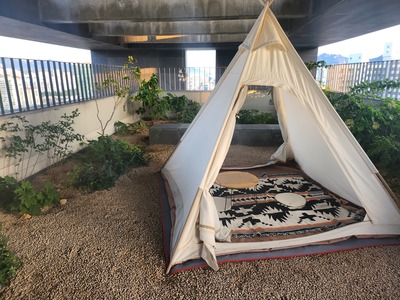}
        \includegraphics[width=.18\linewidth]{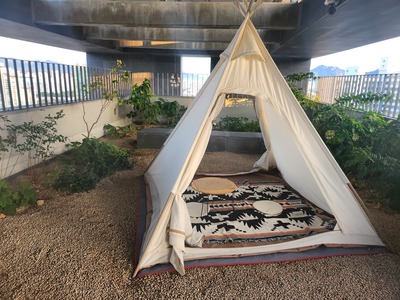}
        \includegraphics[width=.18\linewidth]{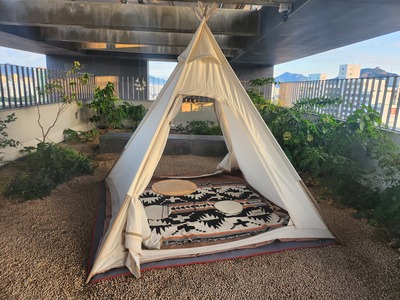}
        \includegraphics[width=.18\linewidth]{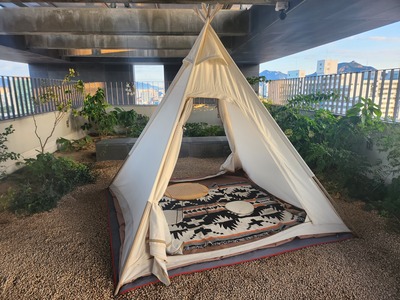}
        \includegraphics[width=.18\linewidth]{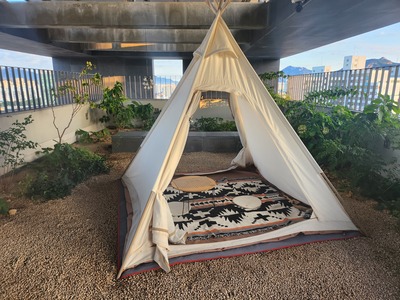}
        \vspace{.125cm}
    \end{subfigure}
    \begin{subfigure}[b]{1.0\textwidth}
    \centering
        \includegraphics[width=.18\linewidth]{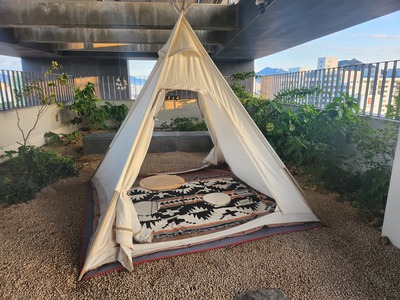}
        \includegraphics[width=.18\linewidth]{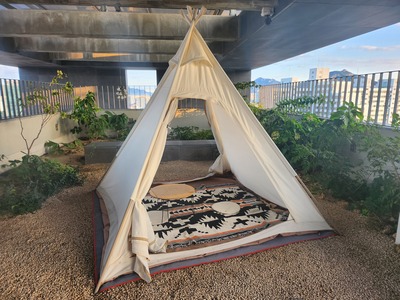}
        \includegraphics[width=.18\linewidth]{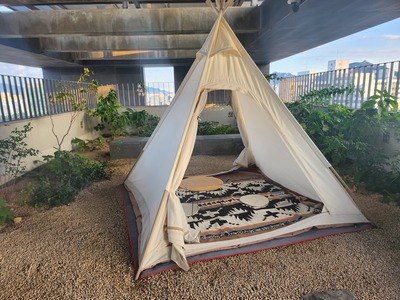}
        \includegraphics[width=.18\linewidth]{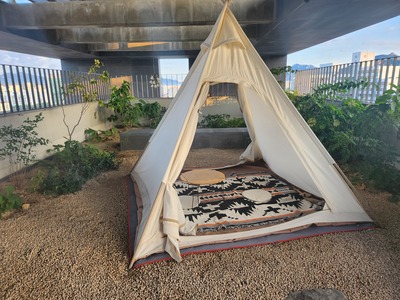}
        \includegraphics[width=.18\linewidth]{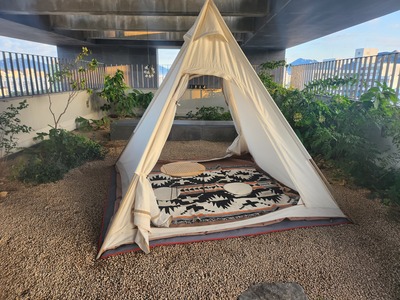}
    \end{subfigure}
    \caption{
        \textbf{Illumination variation samples of the \textit{tent} scene in NeRF Extreme. }}
    \label{fig:tentall} 
\end{figure*}

\begin{figure*}
    \centering
    \begin{subfigure}[b]{1.0\textwidth}
         \centering
        \includegraphics[width=.18\linewidth]{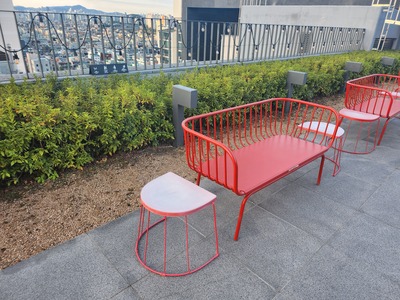}
        \includegraphics[width=.18\linewidth]{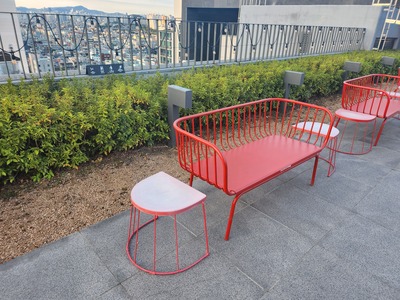}
        \includegraphics[width=.18\linewidth]{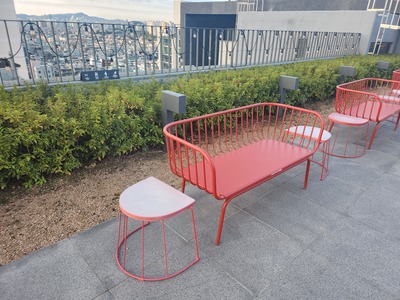}
        \includegraphics[width=.18\linewidth]{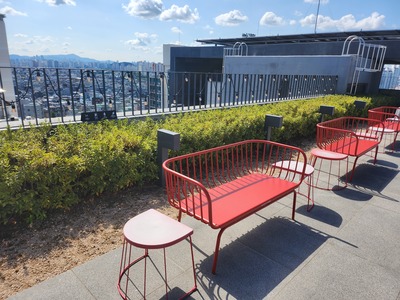}
        \includegraphics[width=.18\linewidth]{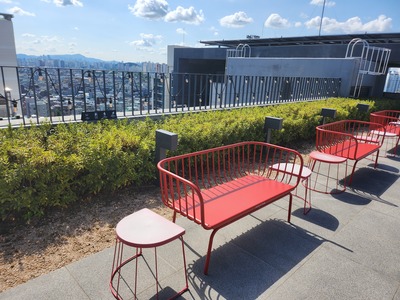}
        \vspace{.125cm}
    \end{subfigure}
    \begin{subfigure}[b]{1.0\textwidth}
    \centering
        \includegraphics[width=.18\linewidth]{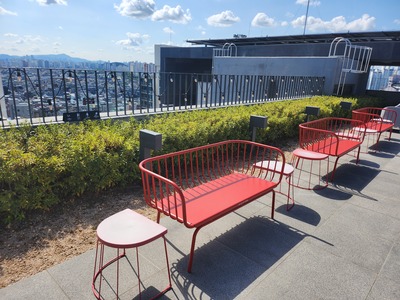}
        \includegraphics[width=.18\linewidth]{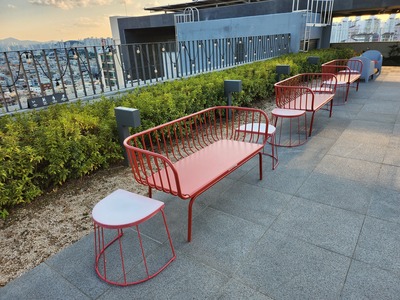}
        \includegraphics[width=.18\linewidth]{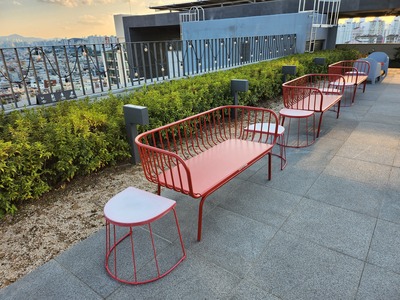}
        \includegraphics[width=.18\linewidth]{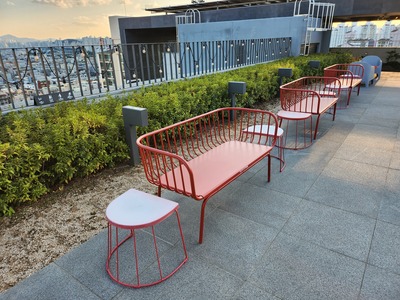}
        \includegraphics[width=.18\linewidth]{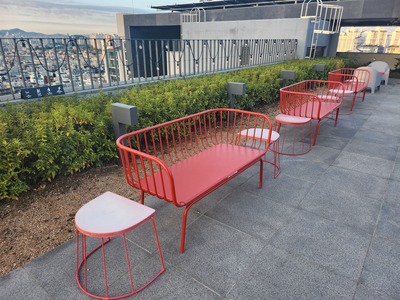}
        \vspace{.125cm}
    \end{subfigure}
    \begin{subfigure}[b]{1.0\textwidth}
    \centering
        \includegraphics[width=.18\linewidth]{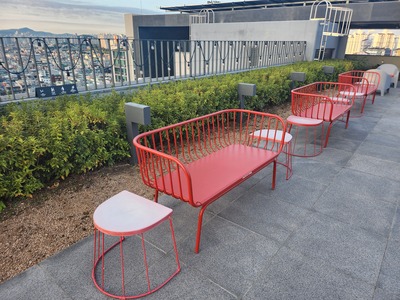}
        \includegraphics[width=.18\linewidth]{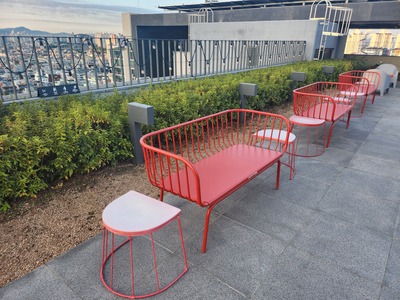}
        \includegraphics[width=.18\linewidth]{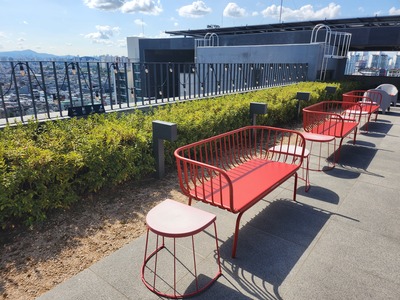}
        \includegraphics[width=.18\linewidth]{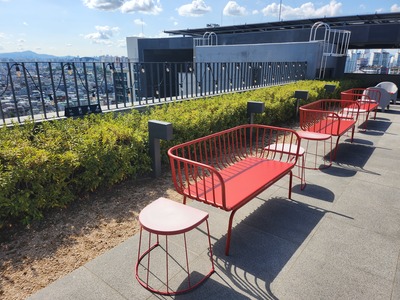}
        \includegraphics[width=.18\linewidth]{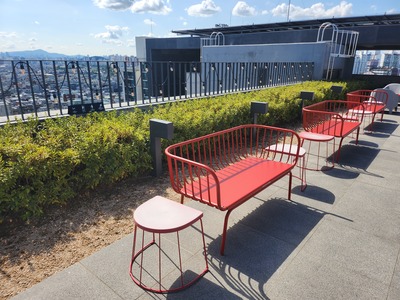}
        \vspace{.125cm}
    \end{subfigure}
    \begin{subfigure}[b]{1.0\textwidth}
    \centering
        \includegraphics[width=.18\linewidth]{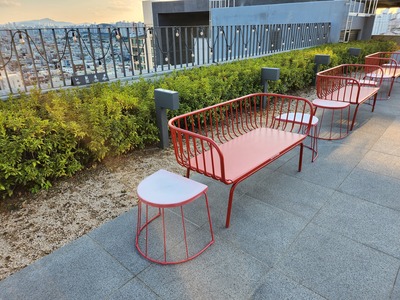}
        \includegraphics[width=.18\linewidth]{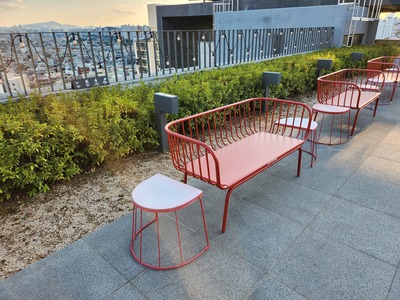}
        \includegraphics[width=.18\linewidth]{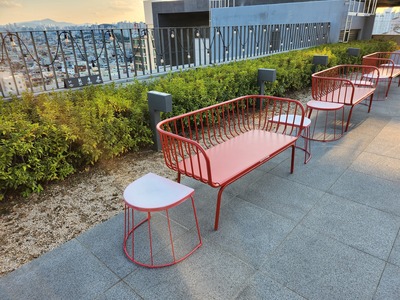}
        \includegraphics[width=.18\linewidth]{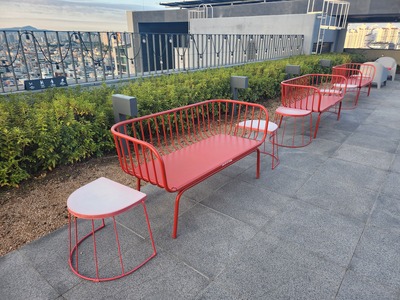}
        \includegraphics[width=.18\linewidth]{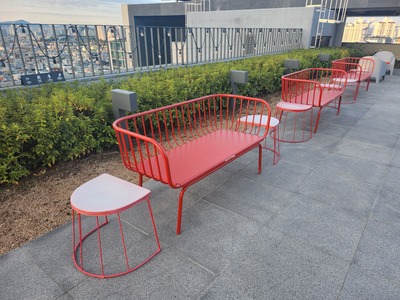}
        \vspace{.125cm}
    \end{subfigure}
    \begin{subfigure}[b]{1.0\textwidth}
    \centering
        \includegraphics[width=.18\linewidth]{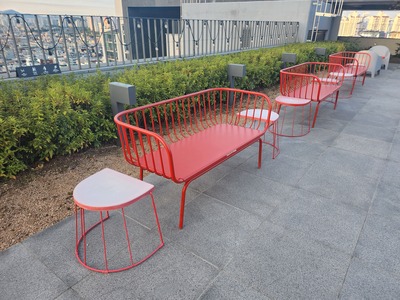}
        \includegraphics[width=.18\linewidth]{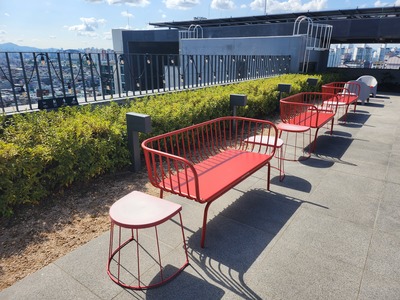}
        \includegraphics[width=.18\linewidth]{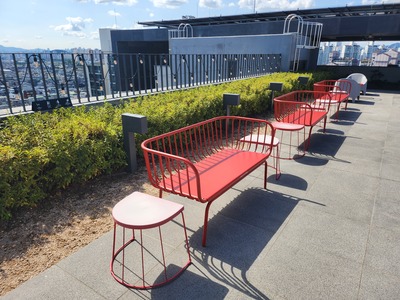}
        \includegraphics[width=.18\linewidth]{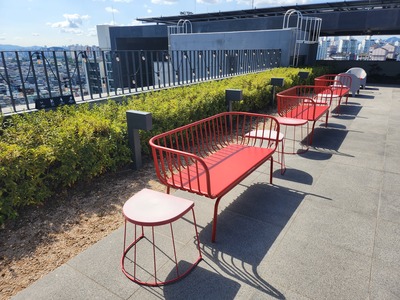}
        \includegraphics[width=.18\linewidth]{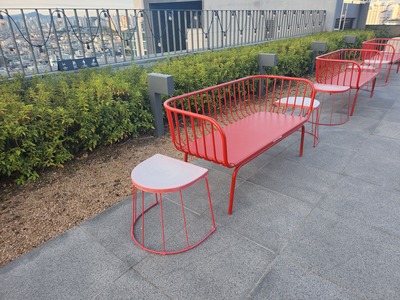}
        \vspace{.125cm}
    \end{subfigure}
    \begin{subfigure}[b]{1.0\textwidth}
    \centering
        \includegraphics[width=.18\linewidth]{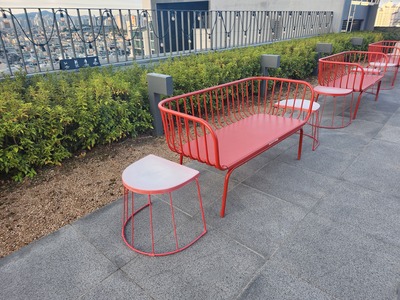}
        \includegraphics[width=.18\linewidth]{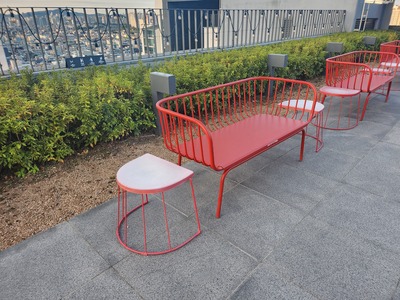}
        \includegraphics[width=.18\linewidth]{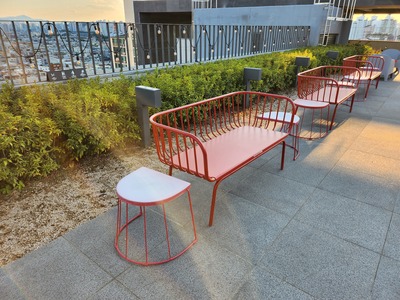}
        \includegraphics[width=.18\linewidth]{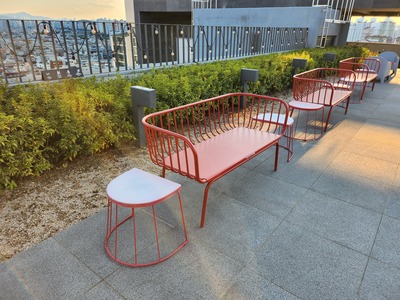}
        \includegraphics[width=.18\linewidth]{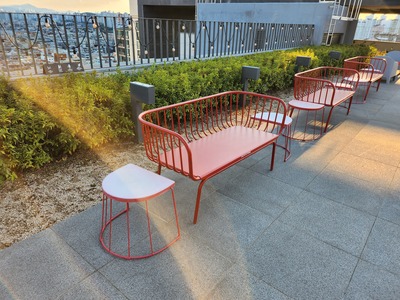}
        \vspace{.125cm}
    \end{subfigure}
    \begin{subfigure}[b]{1.0\textwidth}
    \centering
        \includegraphics[width=.18\linewidth]{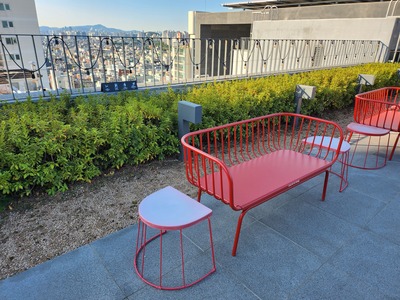}
        \includegraphics[width=.18\linewidth]{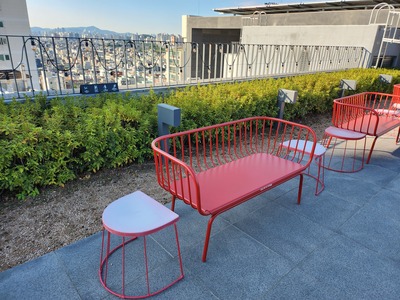}
        \includegraphics[width=.18\linewidth]{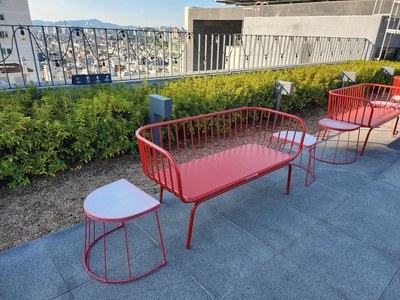}
        \includegraphics[width=.18\linewidth]{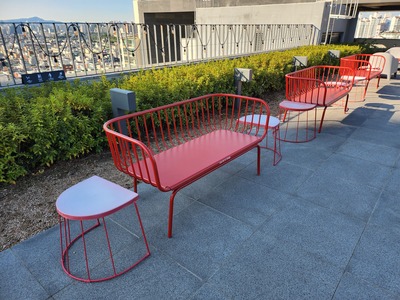}
        \includegraphics[width=.18\linewidth]{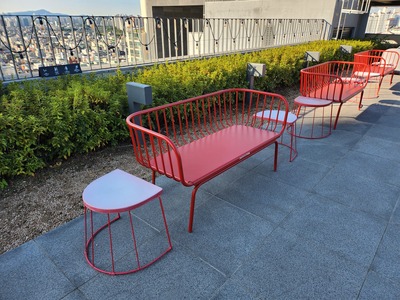}
        \vspace{.125cm}
    \end{subfigure}
    \begin{subfigure}[b]{1.0\textwidth}
    \centering
        \includegraphics[width=.18\linewidth]{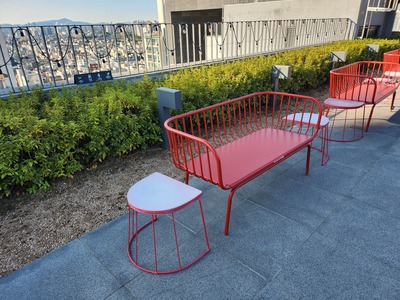}
        \includegraphics[width=.18\linewidth]{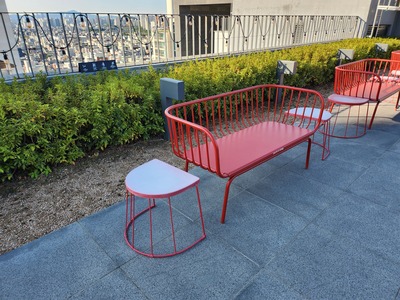}
        \includegraphics[width=.18\linewidth]{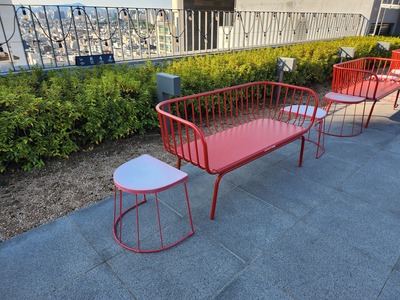}
        \includegraphics[width=.18\linewidth]{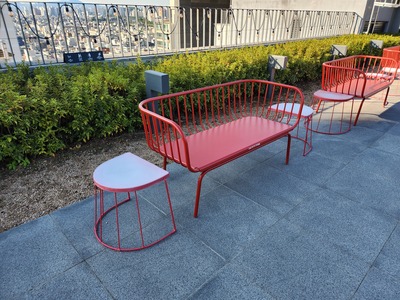}
        \includegraphics[width=.18\linewidth]{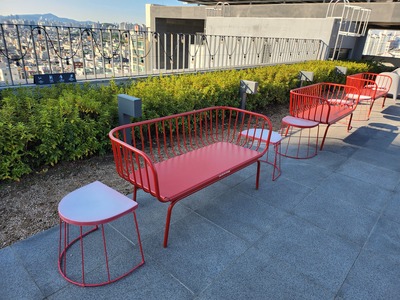}
    \end{subfigure}
    \caption{
        \textbf{Illumination variation samples of the \textit{bench} scene in NeRF Extreme. }}
    \label{fig:benchall} 
\end{figure*}

In this section, we provide exemplified images belonging to our NeRF Extreme benchmark.  Fig.~\ref{fig:tableall} and \ref{fig:roomall} show all the training and test images belonging to our `Table' and `Room' scenes, respectively. Indoor scenes are captured with turned-on/off lights, and closed/open curtains, to get illumination variance. Fig.~\ref{fig:tentall} and \ref{fig:benchall} show all the training and the test images(2 rows from the last) belonging to our  `Tent' and `Bench' scene, respectively. Outdoor scenes are captured at different times and in different sunlight, to be taken under varying illumination conditions.

\begin{table*}
\resizebox{\linewidth}{!}{\begin{tabular}{l|ccc|ccc|ccc|ccc}
\toprule
 & \multicolumn{3}{c|}{Scan 38} & \multicolumn{3}{c|}{Scan 41} & \multicolumn{3}{c}{Scan 82} & \multicolumn{3}{c}{Scan 110} \\ \midrule
  & SSIM $\uparrow$ & LPIPS $\downarrow$ & Abs Rel $\downarrow$ & SSIM $\uparrow$& LPIPS $\downarrow$ & Abs Rel $\downarrow$ & SSIM $\uparrow$& LPIPS $\downarrow$ & Abs Rel $\downarrow$ & SSIM $\uparrow$& LPIPS $\downarrow$ & Abs Rel $\downarrow$\\ \midrule
NeRF-W~\cite{nerfinthewild} \  & 0.17 & 0.75 & 0.43  & 0.22 & 0.69 & 0.65 & 0.41 & 0.55 & 0.69 & 0.36 & 0.59 & 0.84 \\
Ha-NeRF~\cite{HaNeRF} \  & 0.10 & 0.74 & 0.52  & 0.25 & 0.70 & 0.43 & 0.32 & 0.60 & \underline{0.66} & 0.39 & 0.57 & \underline{0.83} \\
\midrule
RegNeRF~\cite{niemeyer2022regnerf}\  & \underline{0.40} & \underline{0.44} & \textbf{0.29} & \textbf{0.51} & \textbf{0.41} & \underline{0.26} & \textbf{0.77} &\textbf{ 0.23} & \textbf{0.42} & \underline{0.63} & \textbf{0.40} & 0.96 \\
FreeNeRF~\cite{yang2023freenerf} \  & 0.22 & 0.59 & \underline{0.42}  & 0.37 & 0.52 & 0.66 & \textbf{0.77} & \underline{0.24} & 0.67 & 0.44 & 0.54 & 2.03 \\
\midrule
NeROIC-Geom.~\cite{kuang2022neroic}\  & 0.22 & 0.52 &\underline{0.42}  & 0.17 & 0.73 & 0.59 & 0.42 & 0.51 & 0.69 & 0.18 & 0.58 & 0.84 \\
\textbf{ExtremeNeRF (Ours)}\  &\textbf{ 0.46} & \textbf{0.41} & \textbf{0.29}  & \underline{0.46} & \underline{0.43} & \textbf{0.24 }& \underline{0.72} & 0.28 & \textbf{0.42} & \textbf{0.64} & \underline{0.43} & \textbf{0.80}\\
\bottomrule
\end{tabular}
}
    \caption{
    \textbf{Additional quantitative comparison on DTU~\cite{DTU}.}
    We compare our quantitative result with the baseline methods in $3$ view settings. In most cases, our proposed method ranks the best or the second-best in performance.
    }
    \label{tab:dtu} 
\end{table*}

\begin{table*}
\resizebox{\linewidth}{!}{\begin{tabular}{l|ccc|ccc|ccc|ccc}
\toprule
 & \multicolumn{3}{c|}{Houseplant} & \multicolumn{3}{c|}{Room} & \multicolumn{3}{c}{Leaves} & \multicolumn{3}{c}{Tree} \\ \midrule
  & SSIM $\uparrow$ & LPIPS $\downarrow$ & Abs Rel $\downarrow$ & SSIM $\uparrow$& LPIPS $\downarrow$ & Abs Rel $\downarrow$ & SSIM $\uparrow$& LPIPS $\downarrow$ & Abs Rel $\downarrow$ & SSIM $\uparrow$& LPIPS $\downarrow$ & Abs Rel $\downarrow$\\ \midrule
NeRF-W~\cite{nerfinthewild} \ & \underline{0.36} &  0.52 & \underline{0.44} & 0.40 & 0.50 & \underline{0.54} & \underline{0.37} & 0.66 & 0.52 & 0.171 & 0.66 & 0.50\\
Ha-NeRF~\cite{HaNeRF} \ & \underline{0.36} &  0.54 & 0.48 & \underline{0.41} & 0.49 & \textbf{0.53} & \textbf{0.38} & 0.61 & 0.36 & \textbf{0.227} & 0.58 & 0.65 \\
\midrule
RegNeRF~\cite{niemeyer2022regnerf}\ & 0.29 & 0.46 & 0.66 & 0.37 & 0.45 & 0.69 & 0.35 & \underline{0.47} & \underline{0.31} & 0.195 & \underline{0.47} & \underline{0.47} \\
FreeNeRF~\cite{yang2023freenerf} \ & 0.33 & \underline{0.42} & 0.82 & 0.40 & \underline{0.41} & 0.87 & 0.34 & \underline{0.47} & 0.81 & 0.215 & \textbf{0.45} & 0.84 \\
\midrule
NeROIC-Geom.~\cite{kuang2022neroic}\ & 0.26 & 0.65 & 0.55 & 0.21 & 0.63 & 0.68 & 0.25 & 0.73 & 0.54 & 0.082 & 0.74 & 0.68\\
\textbf{ExtremeNeRF (Ours)}\ & \textbf{0.37} & \textbf{0.39} & \textbf{0.43} & \textbf{0.45} &\textbf{0.39} & 0.63 & \textbf{0.38} &\textbf{0.46} & \textbf{0.20} & \underline{0.220} & \underline{0.47} & \textbf{0.40}\\
\bottomrule
\end{tabular}
}
    \caption{
    \textbf{Additional quantitative comparison on NeRF Extreme benchmark.}
    }
    \label{tab:supp_extreme} 
\end{table*}

\end{document}